\PassOptionsToPackage{table}{xcolor}
\RequirePackage{fix-cm}
\documentclass{article} % For LaTeX2e
\usepackage{iclr2027_conference,times}

\usepackage{amsmath,amsfonts,bm}

\def\eqref#1{equation~\ref{#1}}
\def\1{\bm{1}}

\DeclareMathAlphabet{\mathsfit}{\encodingdefault}{\sfdefault}{m}{sl}
\SetMathAlphabet{\mathsfit}{bold}{\encodingdefault}{\sfdefault}{bx}{n}

\usepackage{hyperref}
\usepackage{url}

\usepackage{lipsum}
\usepackage{graphicx}
\usepackage{caption}
\usepackage{subcaption}
\usepackage{amsmath, amssymb} % For math notation like \mathbf
\usepackage{tikz}
\usetikzlibrary{arrows.meta, positioning}
\usepackage{booktabs}
\usepackage{multirow}
\usepackage{titletoc}
\usepackage[table]{xcolor}
\usepackage{tabularx}
\usepackage{tcolorbox}
\definecolor{winGreen}{RGB}{209, 242, 214} % Soft background green
\newcommand{\win}[1]{\cellcolor{winGreen}\textbf{#1}}

\usepackage{float}
\usepackage[section]{placeins} % Prevents floats from crossing section boundaries
\usepackage{pifont}
\usepackage{listings}
\usepackage{pgfplots}
\usepgfplotslibrary{groupplots}
\pgfplotsset{compat=1.18}

\definecolor{citecolor}{HTML}{0071BC}
\definecolor{linkcolor}{HTML}{ED1C24}

\title{Twice Upon a Time: Complementary Representations for Time-Series Reasoning}

\newcommand{\authorblock}[2]{%
  \begin{tabular}[t]{c}\textbf{#1}\\ \normalfont #2\end{tabular}}
\author{%
\hspace*{-\tabcolsep}\makebox[\linewidth][c]{%
  \authorblock{Hedi Zisling\thanks{Equal contribution.  Correspondence to  hediz@post.bgu.ac.il}}{Ben-Gurion University}\hspace{3em}%
  \authorblock{Liran Nochumsohn\footnotemark[1]}{Ben-Gurion University}\hspace{3em}%
  \authorblock{Ilan Naiman}{Google}%
}\\[4ex]
\hspace*{-\tabcolsep}\makebox[\linewidth][c]{%
  \authorblock{Mihir Parmar}{Google}\hspace{3em}%
  \authorblock{Omri Azencot}{Ben-Gurion University}%
}
}

\iclrfinalcopy
\begin{document}

\maketitle
\lhead{Preprint}

\begin{abstract}
Time Series Language Models (TSLMs) enable language-based understanding and
reasoning over temporal data, yet a fundamental question remains: what
information should their time-series representations preserve, and what
variation should they factor out? Existing TSLMs typically expose the language
model to a single representation, imposing a fixed inductive bias across
downstream tasks. We instead introduce complementary representations as a
design principle for time-series reasoning. Our model combines two visual
views: line plots preserve absolute measurements and the observed coordinate
frame, while normalized delay embeddings factor out positive amplitude
rescaling and offsets while exposing temporal structure. Dedicated visual
encoders integrate both views into a shared language model through visual
tokens and multi-level feature fusion. Evaluation-time interventions verify
that the two views exhibit distinct invariance profiles, while single-view
ablations show that their combination is particularly beneficial beyond basic
perception, with the large gains in prediction and reasoning. Built on Qwen3-VL-8B, our
dual-view model improves overall TSRBench accuracy from 44.5\% with line plots
alone to 53.2\%, substantially exceeding the strongest evaluated dedicated
TSLM baseline at 36.7\%. It further achieves 89.9\% accuracy on TSExam and
86.8\% numerical accuracy on multivariate ChatTS alignment. Together, these
results demonstrate that complementary representations provide an effective
design principle for general-purpose TSLMs, enabling models to exploit temporal
structure without sacrificing access to the original measurement frame.
\par\vspace{0.2em}
\begin{center}
\textbf{Links:}
\href{https://github.com/azencot-group/dual-time}{\textcolor{magenta}{Code}} (GitHub) \;$|$\;
\href{https://huggingface.co/SequentialLearning/DualTime-8B}{\textcolor{cyan}{Models}} (Hugging Face) \;$|$\;
\href{https://huggingface.co/datasets/SequentialLearning/DualTime-data}{\textcolor{teal}{Data}} (Hugging Face)
\end{center}
\end{abstract}

\section{Introduction}

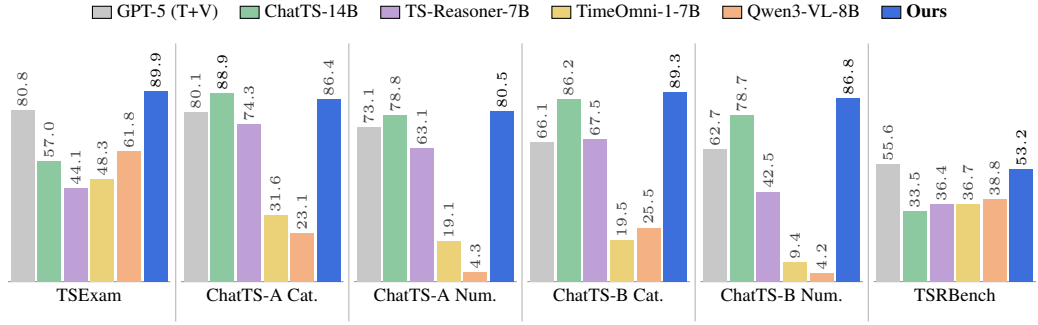
\begin{figure}[t]
\centering
\definecolor{cGPT}{RGB}{200,200,200}
\definecolor{cQwen}{RGB}{244,177,131}
\definecolor{cChatTS}{RGB}{140,200,160}
\definecolor{cTSR}{RGB}{190,160,215}
\definecolor{cTO}{RGB}{235,210,120}
\definecolor{cOurs}{RGB}{60,110,220}
\pgfplotsset{
  cbar/.style={ybar, bar shift=0pt, bar width=0.88, draw=none, fill=#1},
  hl1/.style={}, hl2/.style={}, hl3/.style={},
  hl4/.style={}, hl5/.style={}, hl6/.style={},
}
\newcommand{\benchpanel}[8]{
  \nextgroupplot[title={#1},
    hl#8/.style={every node near coord/.append style={font=\fontsize{5.5}{6}\selectfont\bfseries, text=black}}]
  \addplot[cbar=cGPT,    hl1] coordinates {(1,#2)};
  \addplot[cbar=cChatTS, hl2] coordinates {(2,#3)};
  \addplot[cbar=cTSR,    hl3] coordinates {(3,#4)};
  \addplot[cbar=cTO,     hl4] coordinates {(4,#5)};
  \addplot[cbar=cQwen,   hl5] coordinates {(5,#6)};
  \addplot[cbar=cOurs,   hl6] coordinates {(6,#7)};
}
\begin{tikzpicture}
\begin{groupplot}[
  group style={group size=6 by 1, horizontal sep=4pt, group name=G},
  scale only axis, width=2.15cm, height=2.8cm,
  xmin=0.45, xmax=6.55, ymin=0, ymax=100, clip=false,
  xtick=\empty, hide y axis,
  axis x line*=bottom, axis line style={black!40},
  title style={font=\scriptsize, at={(0.5,0)}, anchor=north, yshift=-5pt},
  nodes near coords={\pgfmathprintnumber[fixed,fixed zerofill,precision=1]{\pgfplotspointmeta}},
  every node near coord/.append style={font=\fontsize{5.5}{6}\selectfont,
      rotate=90, anchor=west, inner sep=1pt, text=black!75},
  legend style={at={(3.164,1.18)}, anchor=south, legend columns=6,
                draw=none, fill=none, font=\scriptsize,
                /tikz/every even column/.append style={column sep=6pt}},
  legend image code/.code={\draw[#1] (0cm,-0.08cm) rectangle (0.25cm,0.08cm);},
]
\benchpanel{TSExam}           {80.8}{57.0}{44.1}{48.3}{61.8}{89.9}{6}
\legend{GPT-5 (T+V), ChatTS-14B, TS-Reasoner-7B, TimeOmni-1-7B, Qwen3-VL-8B, \textbf{Ours}}
\benchpanel{ChatTS-A Cat.}    {80.1}{88.9}{74.3}{31.6}{23.1}{86.4}{2}
\benchpanel{ChatTS-A Num.}    {73.1}{78.8}{63.1}{19.1}{4.3} {80.5}{6}
\benchpanel{ChatTS-B Cat.}    {66.1}{86.2}{67.5}{19.5}{25.5}{89.3}{6}
\benchpanel{ChatTS-B Num.}    {62.7}{78.7}{42.5}{9.4} {4.2} {86.8}{6}
\benchpanel{TSRBench}         {55.6}{33.5}{36.4}{36.7}{38.8}{53.2}{6}
\end{groupplot}
\foreach \i in {1,...,5}{
  \draw[black!30] ([xshift=2pt,yshift=-15pt]G c\i r1.south east)
               -- ([xshift=2pt,yshift=10pt]G c\i r1.north east);
}
\end{tikzpicture}
\caption{Cross-benchmark accuracy (\%). ChatTS-A and ChatTS-B report categorical (Cat.) and numerical (Num.) alignment on the univariate and multivariate datasets, respectively. Bold marks the best result in each panel, excluding GPT-5 (T+V), shown in gray, which is a large proprietary model included as a reference rather than a size-matched comparison.}
\label{fig:cross_benchmark_summary}
\vspace{-2em}
\end{figure}

Time Series Language Models (TSLMs) seek to extend the reasoning and interaction capabilities of large language models to sequential data, enabling models to answer questions, infer temporal relationships, forecast future behavior, and support decisions over time series~\citep{xie2025chatts, langer2026opentslm, guan2026timeomni, yu2026tsrbench}. A central challenge in this emerging paradigm is how temporal signals should be represented to the language model. Existing approaches serialize numerical values into text, learn dedicated time-series embeddings, or project features from specialized temporal encoders into the language-model representation space~\citep{xie2025chatts, langer2026opentslm, guan2026timeomni}. As summarized in Table~\ref{tab:tslm_comparison}, these approaches typically commit to a single time-series representation, with absolute measurement scale either retained directly or provided separately as auxiliary metadata. Yet these representational choices impose fundamentally different inductive biases: representations that retain absolute amplitudes and coordinate information support quantitative readout, whereas normalized representations can suppress such information while emphasizing relative structure and temporal dynamics. Recent evaluations further show that simply rendering a time series as a line plot can already provide strong temporal understanding to general-purpose vision-language models~\citep{yu2026tsrbench, cai2024timeseriesexam}, suggesting that representation---not only model scale or training strategy---is a key determinant of time-series reasoning ability. This motivates a more fundamental question:
    \textit{Can complementary representations improve
general-purpose time-series reasoning?}

We hypothesize that general-purpose time-series reasoning benefits from exposing the model to \emph{complementary views} of the same signal, each preserving different information and inducing different invariances. Rather than requiring a single representation to simultaneously retain all information relevant to downstream reasoning while suppressing nuisance variation, this formulation allows distinct views to provide different inductive biases. We instantiate this principle with two visual representations: a standard line plot and a normalized delay embedding. The line plot \emph{preserves} absolute amplitudes, axis ranges, and measurement scales, providing the information required for questions grounded in observed values. In contrast, delay embeddings reconstruct a time series as a trajectory in phase space using time-lagged observations, revealing geometric structure associated with the underlying dynamics \citep{takens2006detecting}. We normalize this representation before rendering it as a two-dimensional view, thereby \emph{factoring out} affine changes in amplitude while retaining the geometry of the temporal trajectory. Our central hypothesis is that combining these complementary views is particularly useful beyond basic perception, where reasoning, prediction, and decision-making require models to exploit temporal structure while remaining grounded in the observed measurements.

Building on this representational principle, we design a dual-tower TSLM in which each visual view is processed by a dedicated encoder. We retain Qwen3-VL's native vision encoder, initialized from SigLIP-2 \citep{tschannen2025siglip}, to process line plots, and introduce a dedicated DINOv3 encoder \citep{simeoni2025dinov3} for delay embeddings, allowing each view to develop a specialized visual representation before interacting with the language model. The resulting visual features are compressed into a compact set of tokens, 230 for the line plot and 64 for the delay embedding, and concatenated as input to the language model.
Beyond this input-level fusion, we adopt DeepStack \citep{meng2024deepstack} to inject multi-level visual features into the early language-model layers, enabling the complementary representations to interact with language processing throughout the network. 
This architecture provides a compact realization of our central hypothesis while retaining the distinct inductive biases of the two visual views.

To test our representational hypothesis and evaluate the resulting dual-tower TSLM, we conduct experiments on TSRBench~\citep{yu2026tsrbench}, TSExam~\citep{cai2024timeseriesexam}, and ChatTS~\citep{xie2025chatts}. Built on Qwen3-VL-8B~\citep{bai2025qwen3}, our model achieves 53.2\% overall accuracy on TSRBench, substantially narrowing the gap to GPT-5 with text and visual input (55.6\%) while surpassing the strongest evaluated dedicated TSLM (36.7\%). It also outperforms the compared general-purpose and specialized baselines on TSExam and ChatTS numerical alignment. To assess the contribution of the complementary view, we compare against a plot-only variant trained under the same setup. Adding the delay embedding raises overall TSRBench accuracy from 45.2\% to 53.2\%, an 8.0-percentage-point improvement, with gains in reasoning, prediction, and decision-making while leaving perception nearly unchanged. Controlled evaluation-time interventions further reveal distinct invariance profiles: the delay-only model maintains nearly unchanged accuracy under positive rescaling and offset shifts, whereas the plot-only model, whose representation retains the measurement frame, is substantially more sensitive to these transformations. Our contributions are threefold:

\begin{itemize}
    \item \textbf{Complementary representations for time-series reasoning.}
    We establish complementary representation as a design principle for TSLMs, motivated by the observation that quantitative readout requires the original measurement frame, whereas reasoning about temporal dynamics benefits from factoring it out.

    \item \textbf{A dual-view time-series language model.}
    We propose a dual-tower architecture that combines line plots and normalized delay embeddings within a shared language model. Dedicated visual encoders, compact visual tokens, and multi-level feature fusion integrate temporal structure with the original measurement frame.

    \item \textbf{State-of-the-art TSLM performance.}
    With an 8B backbone, our model outperforms all evaluated dedicated TSLMs on TSRBench, achieving 53.2\% accuracy versus 36.7\% for the strongest baseline. It also leads the evaluated models on TSExam and multivariate ChatTS numerical alignment, achieving 89.9\% and 86.8\% accuracy, respectively.
\end{itemize}

\begin{table}[h!]
\centering
\small
\setlength{\tabcolsep}{4.5pt}
\caption{
Comparison of time-series representations in existing TSLMs.
The methods differ in how the time series is represented, whether the
primary representation is normalized, and how absolute measurement-scale
information is provided to the language model.
Forecasting-oriented LLM methods are included for completeness; unlike TSLMs, they target numerical forecasting rather than general question answering and reasoning.}
\label{tab:tslm_comparison}

\resizebox{\columnwidth}{!}{%
\begin{tabular}{llccccc}
\toprule
 & Model &
TS Representation &
Normalization &
Scale Information &
Representation Views &
Task Scope \\
\midrule

\multirow{3}{*}{\rotatebox[origin=c]{90}{\scriptsize Forecast.}}
 & PromptCast \citep{xue2023promptcast}
    & Text
    & None
    & In representation
    & Single
    & Forecasting \\

 & LLMTime \citep{gruver2023large}
    & Text
    & Per-series rescaling
    & Not provided
    & Single
    & Forecasting \\

 & Time-LLM \citep{jin2024timellm}
    & Reprogrammed patches
    & Per-series
    & Not provided
    & Single
    & Forecasting \\

\midrule

\multirow{4}{*}{\rotatebox[origin=c]{90}{\scriptsize TSLMs}}
 & TimeOmni-1 \citep{guan2026timeomni}
    & Text
    & None
    & In representation
    & Single
    & QA \& reasoning \\

 & ChatTS \citep{xie2025chatts}
    & 1D patches
    & Per-series
    & Auxiliary metadata
    & Single
    & QA \& reasoning \\

 & OpenTSLM \citep{langer2026opentslm}
    & 1D patches
    & Per-series
    & Auxiliary metadata
    & Single
    & QA \& reasoning \\

 & TS-Reasoner \citep{yu2026tsreasoner}
    & TSFM features
    & Per-series
    & Auxiliary metadata
    & Single
    & QA \& reasoning \\

\midrule

 & \textbf{Ours}
    & Plot + delay embedding
    & View-dependent
    & Line plot
    & Complementary
    & QA \& reasoning \\

\bottomrule
\end{tabular}%
}
\end{table}

\section{Related Work}

\paragraph{Time-Series Representations.}

Time series admit multiple representations that expose different properties of the underlying signal. Beyond the raw sequential domain, classical approaches transform temporal observations into alternative coordinate systems, including frequency and time-frequency representations, recurrence-based representations, and delay embeddings that reconstruct trajectories in phase space~\citep{takens2006detecting}. A related line of work represents time series as images, enabling spatial models to operate on temporal structure. \cite{wang2015imaging} introduced Gramian Angular Fields and Markov Transition Fields for encoding temporal sequences as two-dimensional images, demonstrating their utility for classification and imputation. More recently, \cite{naiman2024utilizing} showed that image transformations based on delay embeddings and short-time Fourier transforms can provide effective interfaces between time series and modern vision-based diffusion models, supporting generation across both short and long sequences. Collectively, these works demonstrate that alternative representations can expose useful and distinct structure in temporal signals. Our work builds on this perspective but shifts the focus from selecting a representation for a particular predictive or generative task to \emph{combining complementary representations for general-purpose time-series reasoning}: we jointly expose the model to views designed to retain different information and induce different invariances.

\paragraph{Time-Series Language Models.}

A growing body of work investigates how pretrained language models can be adapted to temporal data. Early approaches primarily focused on forecasting, either by serializing numerical observations as text~\citep{gruver2023large}, prompting language models with textual descriptions of temporal sequences~\citep{xue2023promptcast}, or aligning and reprogramming time-series representations to exploit pretrained language-model representations~\citep{zhou2023onefitsall, sun2023test, jin2024timellm}. More recent work has broadened this paradigm from task-specific prediction toward \emph{Time Series Language Models} that accept time series as a modality and support natural-language understanding and reasoning. ChatTS~\citep{xie2025chatts} aligns learned time-series embeddings with an LLM through instruction tuning, while OpenTSLM~\citep{langer2026opentslm} investigates both soft-prompt and cross-attention interfaces for reasoning over multivariate temporal signals. TimeOmni-1~\citep{guan2026timeomni} instead represents numerical sequences directly within the language-model token space and targets complex temporal reasoning, whereas TS-Reasoner~\citep{yu2026tsreasoner} aligns representations from a pretrained time-series foundation model with an LLM. In parallel, benchmark studies have shown that general-purpose vision-language models can reason surprisingly effectively over time series presented as line plots~\citep{cai2024timeseriesexam,yu2026tsrbench}, providing an alternative to explicitly learned time-series interfaces. Despite these diverse strategies, existing TSLMs predominantly expose the language model to a single primary representation of each temporal signal. Our work instead treats representation as a source of complementary inductive biases, jointly providing views that deliberately preserve different information and exhibit different invariances.

\section{Method}

\subsection{Problem Formulation and Overview}
\label{sec:problem_formulation}

\begin{figure}[t]
  \vspace{-1em}
  \centering
  \includegraphics[width=\textwidth]{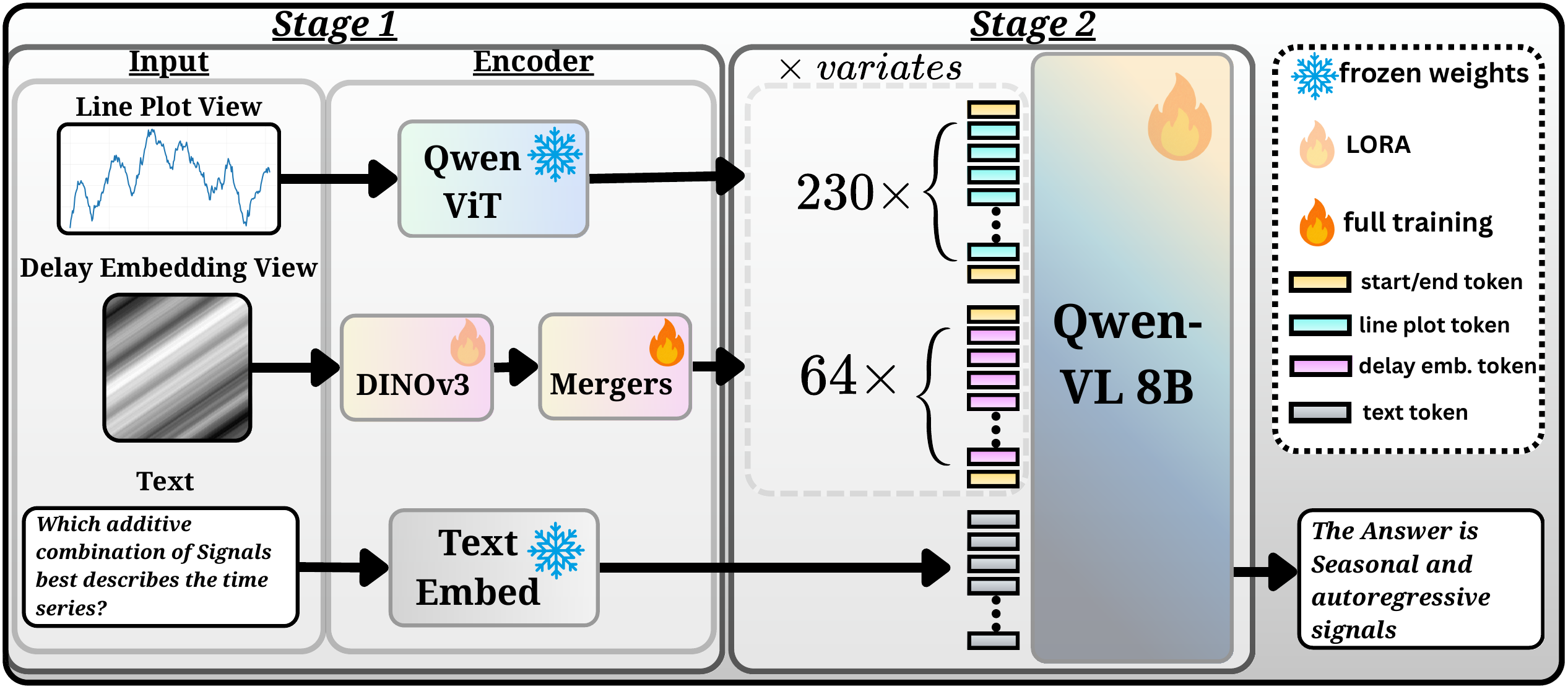}
  \caption{Overview of the Dual-View TSLM. Each visual span is wrapped by \texttt{vision\_start} and \texttt{vision\_end}.
  Stage~I trains the DINOv3 LoRA and the merger.
  Stage~II trains the LM LoRA. The Qwen vision encoder, the text embedding, and the LLM backbone stay frozen.}
  \vspace{-2em}
\label{fig:model_architecture}
\end{figure}

Let $Q$ denote a natural-language query and
$\mathcal{X}=\{\mathbf{x}_1,\ldots,\mathbf{x}_n\}$ a collection of $n$ input time series, where
$\mathbf{x}_i=(x_{i,1},\ldots,x_{i,m_i})\in\mathbb{R}^{m_i}$ may have a different length and numerical range. Given $(Q,\mathcal{X})$, a Time Series
Language Model (TSLM) generates a natural-language answer $A=(a_1,\ldots,a_{|A|})$ autoregressively:
\begin{equation}
P(A\mid Q,\mathcal{X}) = \prod_{j=1}^{|A|}
P(a_j\mid a_{<j},Q,\mathcal{X}) \ .
\end{equation}
Rather than encoding each $\mathbf{x}_i$ through a single representation, we construct two complementary visual views, $\mathcal{V}(\mathbf{x}_i)=\{V_i^{\mathrm{plot}},V_i^{\mathrm{delay}}\}$. The line-plot view $V_i^{\mathrm{plot}}$ retains the original numerical scale and measurement frame, whereas the normalized delay view $V_i^{\mathrm{delay}}$ factors out affine amplitude variation and represents the signal through its local temporal trajectories. As illustrated in Figure~\ref{fig:model_architecture}, the two views are processed by dedicated visual encoders and projected into a shared language-model representation space. Their visual tokens are jointly provided to a single Qwen3-VL-8B language-model backbone, with multi-level visual features additionally injected through DeepStack~\citep{meng2024deepstack}. The model is trained in two stages, first aligning the delay-embedding stream with the language model and subsequently instruction-tuning the language model for time-series reasoning.

\subsection{Complementary Visual Representations}
\label{sec:representations}

Our representation design exposes each time series through two complementary views with different information-preserving properties. The first retains the original measurement frame through a standard line plot, while the second represents the series through overlapping temporal windows after removing its absolute amplitude and offset.

\paragraph{Line-Plot View.} For each time series
$\mathbf{x}_i=(x_{i,1},\ldots,x_{i,m_i})$, we render the raw observations as a two-dimensional line plot without amplitude normalization (Figure~\ref{fig:line_plot_view}). Consequently, the plot retains the original value range and its position within the coordinate system, including the numerical values displayed along the vertical axis. This representation therefore preserves information about absolute amplitudes and differences in scale that would be removed by per-series normalization. For multivariate inputs, each series is rendered as its own plot rather than overlaid, so series with different ranges keep their own measurement frames.

\paragraph{Normalized Delay-Embedding View.}
Delay embeddings~\citep{takens2006detecting} describe a time series through its local histories: overlapping windows, each shifted slightly forward in time, are stacked side by side into an image~\citep{naiman2024utilizing}. Because windows overlap, each feature of the signal traces a diagonal line across the image: periodicity appears as parallel stripes, trends as smooth gradients, and local events as isolated streaks (Figure~\ref{fig:delay_view}). Temporal structure thus becomes visual texture that pretrained vision encoders can readily perceive.

Concretely, for each variate $\mathbf{x}$ of length $m$, we extract $W$ overlapping windows of length $\ell$ with evenly spaced starts covering the whole series. The $c$-th window is the array starting at index $p_c = \operatorname{round}\!\big(c\,\tfrac{m-\ell}{W-1}\big)$:
\begin{equation}
V_c = \big(x_{p_c},\, x_{p_c+1},\, \ldots,\, x_{p_c+\ell-1}\big), \qquad c=0,\ldots,W-1 ,
\label{eq:delay_map}
\end{equation}
where $\ell=\min(\max(\lfloor\rho m\rfloor,\ell_{\min}),\ell_{\max})$ is a fixed fraction $\rho$ of the series length, so windows cover a comparable portion of short and long signals, with bounds $\ell_{\min}$ and $\ell_{\max}$ guarding against extreme lengths. For $m<\ell_{\min}$, we set $\ell=m$. Each window is linearly interpolated to $H$ values, and the $W$ resampled windows form the columns of the min--max normalized delay map $V^{\mathrm{delay}}\in[0,1]^{H\times W}$. Every variate, regardless of its length, is thus represented by an image of the same size (hyperparameters in Appendix~\ref{app:delay_details}).

Since resampling is linear and the map is min--max normalized, any transformation $a\mathbf{x}+b$ with $a>0$ yields the same delay map. The line plot therefore retains the measurement frame, whereas the delay map exposes temporal structure independently of positive affine changes in amplitude.

\subsection{Dual-View Architecture}
\label{sec:architecture}

Our architecture integrates the two views into a single Qwen3-VL-8B language-model backbone~\citep{bai2025qwen3} through dedicated visual encoders (Figure~\ref{fig:model_architecture}).

\paragraph{Line-Plot Encoder.} Line plots are processed by the native Qwen3-VL vision pipeline without modification, which merges spatial features over $2\times2$ neighboring patches and projects them to the language-model hidden dimension. This yields $230$ plot tokens $\mathbf{Z}_i^{\mathrm{plot}}$ per series, together with three intermediate feature maps for DeepStack fusion. The encoder remains frozen throughout training.

\paragraph{Delay-Embedding Encoder.} For each time-series variate, we process its normalized delay map $V^{\mathrm{delay}}$ with a dedicated DINOv3~\citep{simeoni2025dinov3} ViT-L/16 encoder denoted by $E_{\mathrm{delay}}$. Its spatial patch features are merged in $2\times2$ groups and mapped to the language-model hidden dimension by a two-layer projection module $M^{(0)}$, yielding $64$ delay tokens $\mathbf{Z}^{\mathrm{delay}} = M^{(0)}\big(E_{\mathrm{delay}}(V^{\mathrm{delay}})\big)$ per variate. To match the multi-level DeepStack interface of the plot stream, we additionally extract intermediate features $E^{(k_i)}_{\mathrm{delay}}(V^{\mathrm{delay}})$ after blocks $k_i$, where $(k_1,k_2,k_3)=(6,12,18)$, and merge and project each with a dedicated module $M^{(i)}$ of the same form as $M^{(0)}$, giving multi-level delay features $\mathbf{D}_i = M^{(i)}\big(E^{(k_i)}_{\mathrm{delay}}(V^{\mathrm{delay}})\big)$, $i\in\{1,2,3\}$, aligned one-to-one with the $64$ delay tokens and injected into the language model as described below.

\paragraph{Dual-View Fusion.} Both streams enter the language model in the same way. For each time series, $\mathbf{Z}_i^{\mathrm{plot}}$ and $\mathbf{Z}_i^{\mathrm{delay}}$ occupy separate visual spans, each framed by \texttt{<|vision\_start|>} and \texttt{<|vision\_end|>} tokens, alongside the textual query $Q$, so the two views interact within a single autoregressive language model rather than through separate heads or an ensemble. Beyond this input-level fusion, we extend DeepStack to the delay stream. Following the DeepStack mechanism of Qwen3-VL~\citep{meng2024deepstack, bai2025qwen3}, the three multi-level features of each stream are added directly to the hidden states of the first three language-model layers, one feature per layer, at the positions of that stream's own visual tokens. The plot and delay streams thus inject into the same layers but at disjoint token positions. Rendering and tokenization details are provided in Appendix~\ref{app:implementation}.

\subsection{Training}
\label{sec:training}

Following the staged training paradigm common in multimodal language models and recent TSLMs~\citep{xie2025chatts, langer2026opentslm}, we train our model, initialized from Qwen3-VL-8B~\citep{bai2025qwen3}, in two stages that separate the alignment of the new delay stream from task-level instruction tuning. In \textbf{Stage~I (modality alignment)}, the Qwen3-VL vision encoder and the language model remain frozen, and we train LoRA adapters~\citep{hu2022lora} on DINOv3 together with the projection $M^{(0)}$, which maps the final DINOv3 patch features to delay tokens, and the projections $\{M^{(i)}\}_{i=1}^{3}$, which map the intermediate DINOv3 features to the multi-level DeepStack features. This stage is trained on captioning data, encouraging the delay stream to produce representations that the frozen language model can interpret. In \textbf{Stage~II (instruction tuning)}, the DINOv3 adapters are merged, the entire visual pathway is frozen, and LoRA is applied to the language model, which is trained on time-series question answering to reason jointly over both views. Both stages use the standard next-token prediction loss computed over the answer tokens only. The objective and hyperparameters are given in Appendix~\ref{app:training_details}.

\section{Training Data}
\label{sec:training_data}

We construct a dedicated training corpus for each of the two stages described in Section~\ref{sec:training}. Stage~I uses caption supervision to align the new delay-embedding stream with the language model, whereas Stage~II uses a heterogeneous instruction corpus to train perception, numerical readout, forecasting, and temporal reasoning. Most time series are generated procedurally, with labels obtained directly from generator parameters or exact functions of the realized signal rather than from a teacher model. The only real-world time-series sources are CaTS-Bench~\citep{zhou-etal-2026-cats} in Stage~I and a small subset of sports win-probability series from GAMETime~\citep{tan2025inferring} used for abductive reasoning in Stage~II. Table~\ref{tab:training_data} summarizes the data composition.

\begin{table}[tbp]
\centering
\small
\caption{Training-data composition. Rows denote distinct examples available
to the corresponding training stage. Stage~I sources are sampled with equal
probability; Stage~II uses a balanced mixture described in the text.}
\label{tab:training_data}
\resizebox{\columnwidth}{!}{%
\begin{tabular}{l l l r l}
\toprule
Stage & Source & Origin & Rows & Supervision \\
\midrule
I & CaTS-Bench & Real~\citep{zhou-etal-2026-cats}
  & 15,995 & Caption \\
I & ChatTS-style captions & Synthetic~\citep{xie2025chatts}
  & 32,000 & Caption \\
I & Structural captions & Synthetic, in-house
  & 7,701 & Caption \\
\midrule
II & ChatTS \texttt{ift} & Synthetic~\citep{xie2025chatts}
   & 6,373 & Free-form QA \\
II & ChatTS \texttt{sft} & Synthetic~\citep{xie2025chatts}
   & 44,802 & Free-form QA \\
II & TSExam perception & Synthetic, in-house
   & 20,640 & MCQ \\
II & Numerical corpus & Synthetic, in-house
   & 11,000 & Scalar \\
II & Reasoning corpus & Synthetic + real sports
   & 100,000 & MCQ / rationale+MCQ \\
\bottomrule
\end{tabular}%
}
\end{table}

\paragraph{Stage I: Caption Alignment.} The alignment corpus contains 55,696 examples from three sources. CaTS-Bench~\citep{zhou-etal-2026-cats} contributes real-world short time series from diverse physical and socioeconomic domains together with reference descriptions. To provide broader attribute coverage, we additionally construct 32,000 ChatTS-style captions from the ChatTS time-series generator~\citep{xie2025chatts}. Trend, seasonality, noise, and local events are sampled from the generator's attribute space, and captions are rendered directly from those attributes. Finally, we construct 7,701 structural captions by generating series with our TimeSeriesExam fork~\citep{cai2024timeseriesexam} and deterministically rendering one-sentence descriptions from the underlying generative components, parameters, noise processes, and injected anomalies. The three sources are
sampled with equal probability during Stage~I, balancing real-world diversity, free-form synthetic descriptions, and tightly controlled constructive
supervision.

\paragraph{Stage II: Instruction Tuning.} The instruction corpus contains 182,815 examples spanning complementary task families. We include the released ChatTS \texttt{ift} and \texttt{sft} training sets~\citep{xie2025chatts} for open-ended univariate and multivariate question answering. Using our TimeSeriesExam fork, we further generate the TSExam perception set, 20,640 multiple-choice questions covering pattern recognition, anomaly detection, similarity, noise, and causality, together with an 11,000-example numerical corpus of open-ended questions whose answers are exact statistics or dynamical quantities computed from the
realized series. The largest component is an in-house 100,000-example reasoning corpus designed to develop broad reasoning capabilities over time series. It spans numerical, deductive, inductive, abductive, causal, etiological, temporal-relation, forecasting, and event-prediction tasks, with examples generated from procedural rules, dynamical systems, and domain-specific simulators. Because instruction-tuned models are sensitive to the task interface, we align the
mixture of reasoning tasks and their question--answer formats with the capabilities evaluated by TSRBench~\citep{yu2026tsrbench}; the underlying
time series, questions, and answers are generated independently and no benchmark instances are used for training. For tasks requiring explicit derivation, supervision additionally includes deterministic rationales constructed from the simulator state, while the remaining tasks use direct multiple-choice supervision.

\paragraph{Gold-by-Construction Supervision and Benchmark Separation.}
For all in-house synthetic data, supervision is derived from known generative
parameters or exact computations on the realized signal. Captions describe
the sampled components, multiple-choice labels follow directly from the
generating mechanism, numerical targets are exact signal reductions, and
reasoning rationales are rendered from structured intermediate facts. This
design avoids using an external language model to infer labels for the
synthetic data. Importantly, no TSRBench or TimeSeriesExam evaluation examples
are included in training.
The TSExam perception set, structural captions, and numerical corpus are generated with our fork of the public generator and contain freshly sampled series whose lengths differ from those in the benchmark.
Similarly, although the reasoning corpus covers the reasoning task families
and question--answer formats evaluated by TSRBench, all training instances
are generated independently and no benchmark examples are used for training.
We additionally verified the reasoning corpus against the benchmark for exact
series overlap, shared sports-game instances, and exact question--answer
matches. ChatTS evaluation data are likewise excluded; only the released
training splits are used. Additional generator details, per-task statistics,
target formats, and held-out splits are provided in
Appendix~\ref{app:data}.

\section{Experiments and Results}
\label{sec:experiments}

\subsection{Experimental Setup}
\label{sec:benchmarks}

We evaluate on three benchmarks that together probe time-series understanding from low-level perception to high-level reasoning. \textbf{TSRBench} \citep{yu2026tsrbench}, our primary benchmark, comprises 4,125 multiple-choice questions over 15 tasks spanning Perception, Reasoning, Prediction, and Decision-Making. \textbf{TSExam} \citep{cai2024timeseriesexam} contains 746 multiple-choice questions on pattern recognition, noise understanding, anomaly detection, similarity, and causality, and \textbf{ChatTS} \citep{xie2025chatts} measures both fine-grained categorical and numerical alignment and higher-level reasoning over univariate and multivariate series. We compare against two families of models. General-purpose vision-language models receive the series as line plots: Qwen3-VL-8B~\citep{bai2025qwen3}, our own backbone, isolates the effect of our design and training, while GPT-5 serves as a substantially larger proprietary reference rather than a size-matched competitor. We further compare against representative dedicated TSLMs: ChatTS-14B~\citep{xie2025chatts}, TS-Reasoner-7B~\citep{yu2026tsreasoner}, and TimeOmni-1-7B~\citep{guan2026timeomni}. Table~\ref{tab:tsrbench_main} reports task-level TSRBench accuracy together with our representation ablations, 
and Figure~\ref{fig:cross_benchmark_summary} summarizes performance across all three benchmarks.
Complete task-level results and evaluation procedures, including the source of each baseline result, are provided in Appendix~\ref{app:full_results}.

\begin{table*}[t]
\centering
\setlength{\tabcolsep}{2.6pt}
\renewcommand{\arraystretch}{1.08}
\caption{
TSRBench accuracy (\%) across all 15 tasks. Rows are grouped into general-purpose models, dedicated TSLMs, and representation ablations. In the Overall column, \textbf{bold} and \underline{underline} mark the best and second-best results, respectively. Gray rows denote large proprietary models included as reference points rather than size-matched comparisons and are excluded from this ranking. Full task descriptions and results are provided in Appendix~\ref{app:full_results}.
}
\label{tab:tsrbench_main}
\resizebox{\textwidth}{!}{%
\begin{tabular}{c l cccc ccccccc cc cc c}
\toprule
& & \multicolumn{4}{c}{\textbf{Perception}} & \multicolumn{7}{c}{\textbf{Reasoning}} & \multicolumn{2}{c}{\textbf{Prediction}} & \multicolumn{2}{c}{\textbf{Decision}} & \\
\cmidrule(lr){3-6}\cmidrule(lr){7-13}\cmidrule(lr){14-15}\cmidrule(lr){16-17}
& Model & PR & NU & AD & CA & ER & CD & AR & TR & NR & DR & IR & TSF & EP & QuaDM & QuanDM & \textbf{Overall} \\
\midrule
\multirow{3}{*}{\rotatebox[origin=c]{90}{\textit{General}}} & \cellcolor{black!6} GPT-5 (V) & \cellcolor{black!6} 83.6 & \cellcolor{black!6} 72.4 & \cellcolor{black!6} 69.8 & \cellcolor{black!6} 73.5 & \cellcolor{black!6} 34.6 & \cellcolor{black!6} 37.0 & \cellcolor{black!6} 81.3 & \cellcolor{black!6} 68.1 & \cellcolor{black!6} 64.0 & \cellcolor{black!6} 48.0 & \cellcolor{black!6} 57.0 & \cellcolor{black!6} 38.8 & \cellcolor{black!6} 71.9 & \cellcolor{black!6} 26.9 & \cellcolor{black!6} 30.0 & \cellcolor{black!6} 52.4 \\
& \cellcolor{black!6} GPT-5 (T+V) & \cellcolor{black!6} 84.9 & \cellcolor{black!6} 72.4 & \cellcolor{black!6} 61.2 & \cellcolor{black!6} 70.8 & \cellcolor{black!6} 38.9 & \cellcolor{black!6} 50.3 & \cellcolor{black!6} 93.3 & \cellcolor{black!6} 68.8 & \cellcolor{black!6} 72.0 & \cellcolor{black!6} 44.0 & \cellcolor{black!6} 58.0 & \cellcolor{black!6} 37.8 & \cellcolor{black!6} 78.1 & \cellcolor{black!6} 31.9 & \cellcolor{black!6} 34.3 & \cellcolor{black!6} 55.6 \\
& Qwen3-VL-8B (V) & 54.7 & 57.5 & 62.8 & 65.5 & 24.3 & 31.7 & 52.0 & 23.1 & 33.2 & 19.2 & 29.0 & 51.7 & 30.6 & 36.1 & 29.0 & 38.9 \\
\midrule
\multirow{3}{*}{\rotatebox[origin=c]{90}{\textit{TSLMs}}} & ChatTS-14B & 50.7 & 50.6 & 46.5 & 55.8 & 21.7 & 34.3 & 51.3 & 24.4 & 30.5 & 23.6 & 25.0 & 19.2 & 52.2 & 29.3 & 27.7 & 33.5 \\
& TS-Reasoner-7B & 53.1 & 56.3 & 48.1 & 50.4 & 28.9 & 24.3 & 57.3 & 26.9 & 37.2 & 23.6 & 24.0 & 31.7 & 55.0 & 32.8 & 21.7 & 36.4 \\
& TimeOmni-1-7B & 55.0 & 59.8 & 41.9 & 64.6 & 28.0 & 35.3 & 46.7 & 24.4 & 31.5 & 22.8 & 34.0 & 30.0 & 49.4 & 34.0 & 30.3 & 36.7 \\
\midrule
\multirow{4}{*}{\rotatebox[origin=c]{90}{\textit{Ablations}}} & Delay only & 57.4 & 65.5 & 69.8 & 76.1 & 30.3 & 51.7 & 55.3 & 24.4 & 38.5 & 26.8 & 21.0 & 49.0 & 20.3 & 34.9 & 19.3 & 40.5 \\
& Plot only & 87.3 & 94.3 & 77.5 & 93.8 & 29.7 & 52.0 & 57.3 & 18.1 & 36.5 & 27.2 & 23.0 & 46.2 & 21.7 & 39.1 & 33.3 & \underline{45.2} \\
& Delay+Delay & 67.9 & 62.1 & 79.1 & 82.3 & 32.3 & 40.3 & 64.7 & 34.4 & 40.2 & 26.8 & 19.0 & 54.6 & 15.8 & 33.1 & 21.0 & 42.6 \\
& Plot+Plot & 84.1 & 94.3 & 80.6 & 87.6 & 40.0 & 43.0 & 60.7 & 24.4 & 32.2 & 26.4 & 27.0 & 46.2 & 28.6 & 32.8 & 28.3 & 44.8 \\
\midrule
& \textbf{Ours} & 85.4 & 94.3 & 78.3 & 92.9 & 43.1 & 51.7 & 56.0 & 28.7 & 35.5 & 29.2 & 34.0 & 50.4 & 84.2 & 32.8 & 43.0 & \textbf{53.2} \\
\bottomrule
\end{tabular} %
}
\vspace{-1em}
\end{table*}

\subsection{Do Complementary Representations Improve Temporal Reasoning?}
\label{sec:representation_analysis}

Our central hypothesis is that exposing the language model to complementary representations provides benefits that cannot be obtained from either view alone. We therefore compare the full dual-view model against \textit{Plot-only} and \textit{Delay-only} variants, as well as two same-view controls that retain the complete dual-tower architecture but feed both towers the same representation. In \textit{Plot+Plot}, the DINOv3 tower receives the line plot in place of the delay map; in \textit{Delay+Delay}, the Qwen3-VL tower receives the delay map in place of the line plot. Both controls keep the same encoders, adapters, per-tower token budgets, and two-stage training procedure as our model, so any difference from the full model reflects whether the two streams carry distinct or identical views, rather than differences in architecture or capacity. Table~\ref{tab:tsrbench_main} reports results across all 15 TSRBench tasks, together with representative dedicated TSLMs and general-purpose models.

The results show that the benefit of complementary representations is concentrated beyond basic perception. Plot only already achieves 87.4\% on Perception, and adding the delay view leaves this essentially unchanged at 86.4\%. In contrast, the dual-view model improves Reasoning from 35.8\% to 40.1\%, Prediction from 38.0\% to 61.7\%, and Decision-Making from 36.4\% to 37.6\%, raising overall accuracy from 45.2\% to \textbf{53.2\%}. The task-level results further show that the contribution of the complementary
view varies across tasks, rather than producing a uniform improvement across the benchmark. Importantly, simply duplicating a view does not reproduce the gain: Plot+Plot reaches 44.8\% overall and Delay+Delay reaches 42.6\%, compared with our 53.2\%. This suggests that the improvement is not explained merely by providing a second visual stream, but is associated with combining representations that preserve different properties of the underlying signal.

Our model also exceeds the evaluated dedicated TSLMs, whose strongest overall TSRBench result is 36.7\%. For context, Table~\ref{tab:tsrbench_main} additionally reports GPT-5 results; these models are substantially larger proprietary systems and are therefore not size-matched comparisons. They nevertheless provide a useful reference point: our model reaches 53.2\% overall, surpassing GPT-5 with visual input (52.4\%) and approaching GPT-5 with text and visual input (55.6\%), while exhibiting a markedly different task-level performance profile.

\subsection{How Do the Representations Respond to Distribution Shifts?}
\label{sec:robustness_analysis}

To directly probe the inductive biases induced by the visual representations, we apply controlled transformations to TSRBench inputs at evaluation time. We consider affine changes to amplitude and offset, observational noise and smoothing, and changes in sampling horizon. Tasks whose gold targets depend intrinsically on absolute coordinate values or sampling rates are excluded from this analysis, leaving 2{,}020 items across invariant reasoning, perception, and event prediction tasks. Table~\ref{tab:tsrbench_interventions} reports the resulting accuracy under each transformation, together with clean performance for reference.

The interventions reveal distinct rather than uniformly superior robustness profiles. Consistent with its per-series normalization, the delay-only representation is nearly invariant to the measured affine transformations:
its clean accuracy of 44.7\% changes only slightly under rescaling and offset shifts. Our model remains sensitive to affine transformations because its
line-plot stream continues to encode the observed measurement frame, while maintaining substantially higher absolute accuracy than Plot only across all
reported interventions. Its sensitivity also depends on the type of shift: noise and smoothing degrade both Plot only and our model, whereas our model
changes relatively little under changes in sampling horizon. Thus, combining the two views does not impose a single invariance
profile or provide uniform robustness to every perturbation. Instead, the results support the intended representational distinction: the views exhibit
different sensitivities, allowing the model to retain access to the original measurement frame while also incorporating a representation that factors out
selected amplitude variation.

\begin{table*}[t]
\centering
\small
\setlength{\tabcolsep}{3pt}
\newcolumntype{Y}{>{\centering\arraybackslash}X}
\caption{
Evaluation-time interventions on scale- and sampling-invariant TSRBench subsets (2{,}020 items; excludes NR, DR, IR, TSF, and decision tasks). Each entry reports accuracy (\%).
}
\label{tab:tsrbench_interventions}
\begin{tabularx}{\textwidth}{@{}l Y YYYY c YYY c YY}
\toprule
& & \multicolumn{5}{c}{\textbf{Affine}}
& \multicolumn{4}{c}{\textbf{Noise / Smoothing}}
& \multicolumn{2}{c}{\textbf{Sampling}} \\
\cmidrule(lr){3-7}
\cmidrule(lr){8-11}
\cmidrule(lr){12-13}
Model & Clean
& $\times 0.1$ & $\times 10$ & $+5\sigma$ & $-5\sigma$ & $z$-score
& 20 dB & 10 dB & 0 dB & Smooth
& 256 & 2048 \\
\midrule
Plot only
& 52.7 & 58.2 & 49.6 & 48.9 & 49.2 & 47.5
& 50.6 & 47.4 & 43.2 & 50.1
& 51.6 & 52.9 \\
Delay only
& 44.7 & 43.4 & 44.9 & 44.9 & 44.1 & 44.3
& 42.3 & 38.3 & 35.3 & 40.7
& 45.1 & 42.8 \\
\midrule
\textbf{Ours}
& \win{66.5} & \win{63.4} & \win{64.8} & \win{62.9} & \win{65.0} & \win{60.7}
& \win{65.4} & \win{62.4} & \win{56.6} & \win{60.7}
& \win{64.8} & \win{66.8} \\
\bottomrule
\end{tabularx}
\vspace{-2em}
\end{table*}

\subsection{Cross-Benchmark Generalization}
\label{sec:cross_benchmark}

Finally, we evaluate whether the gains observed on TSRBench transfer to benchmarks emphasizing different aspects of time-series understanding. Figure~\ref{fig:cross_benchmark_summary} summarizes overall TSExam accuracy, categorical and numerical alignment on the univariate (A) and multivariate (B) ChatTS datasets, and TSRBench overall accuracy; per-task results and ChatTS reasoning evaluations are in Appendix~\ref{app:full_results}.
Our model attains 89.9\% accuracy on TSExam, exceeding the compared general-purpose and dedicated TSLM baselines. On ChatTS, it achieves 80.5\% numerical accuracy on univariate sequences and 86.8\% on multivariate sequences, with the largest advantage appearing in the multivariate setting. Together with the TSRBench results, these evaluations indicate that the benefit of the dual representation is not confined to a single benchmark: the model remains strong across perceptual, numerical, and broader temporal reasoning tasks, despite using a Qwen3-VL-8B backbone.

\section{Discussion}
\label{sec:discussion}

Time-series reasoning requires both numerical readout and understanding of temporal dynamics, yet these capabilities favor different representational properties. This tension motivated our central question: can complementary representations improve general-purpose time-series reasoning? We establish complementary representation as a design principle for TSLMs and instantiate it in a dual-view model that integrates measurement-preserving line plots and normalized delay embeddings through dedicated visual encoders and a shared language model. The resulting model raises TSRBench accuracy from 45.2\% for the plot-only variant to 53.2\%, surpassing all evaluated dedicated TSLMs. The gains are concentrated beyond basic perception and are not reproduced when both encoders receive line plots or both receive delay embeddings. The model also leads the evaluated baselines on TSExam and multivariate ChatTS numerical alignment, achieving 89.9\% and 86.8\% accuracy, respectively, demonstrating strong performance across perceptual, numerical, and broader temporal reasoning tasks.

The STFT comparison in Appendix \ref{app:alternative_representation} further shows that the preferred secondary representation varies across evaluations: delay embeddings perform better on TSRBench, while STFT provides small advantages on ChatTS numerical alignment and average reasoning accuracy. This suggests that representation choice should reflect the capabilities a model must support. The intervention analysis likewise distinguishes access to an invariant view from invariance of the combined model, which retains both normalized structure and absolute measurements. Future work could explore additional representations and encoder combinations, including more than two views and query-dependent mechanisms for combining them.

\bibliography{refs}
\bibliographystyle{iclr2027_conference}

\clearpage

\titlecontents{section}[1.5em]
  {\vspace{0.5em}\bfseries\color{black}}
  {\contentslabel{1.5em}}
  {\hspace*{-1.5em}}
  {\titlerule*[0.5pc]{.}\contentspage}

\titlecontents{subsection}[3.8em]
  {\normalfont\color{black}}
  {\contentslabel{2.3em}}
  {\hspace*{-2.3em}}
  {\titlerule*[0.5pc]{.}\contentspage}

\vspace{2em}
\noindent{\Large\textbf{\textsf{Table of Contents for Appendix}}}\par
\vspace{1em}

\begingroup
\hypersetup{linkcolor=black}
\startcontents[appendix]
\printcontents[appendix]{}{1}{\setcounter{tocdepth}{2}}
\endgroup

\clearpage
\appendix
\section*{Appendix}

\section{Time Series Representations}

\subsection{Visualization of Complementary Representations}
\label{app:representation_visualization}

\begin{figure}[h!]
  \centering
  \captionsetup[subfigure]{skip=2pt}%
  \setlength{\fboxsep}{0pt}\setlength{\fboxrule}{0.4pt}%
  \newlength{\dvh}\setlength{\dvh}{0.294\textwidth}% line-plot image height
  \begin{subfigure}[b]{2.444\dvh}
    \centering
    \includegraphics[height=\dvh]{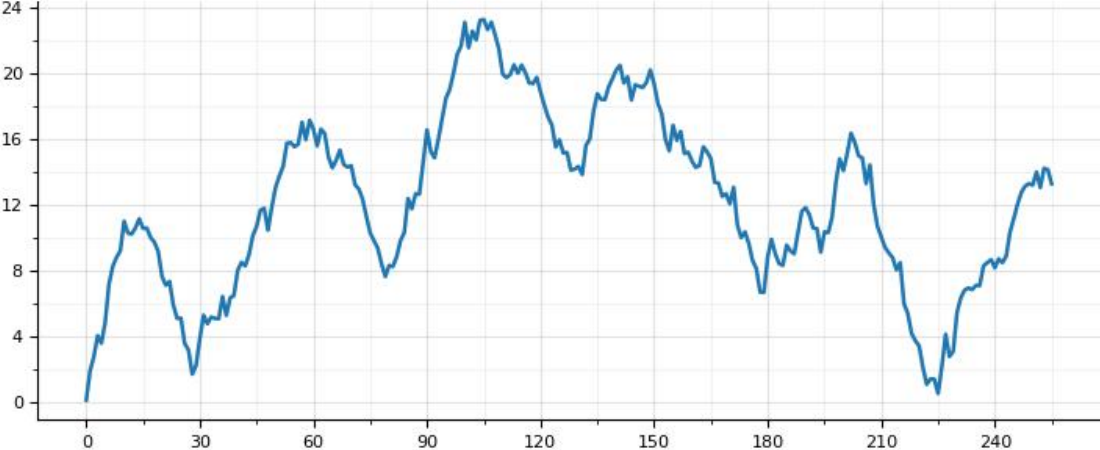}
    \caption{Line-plot view}
    \label{fig:line_plot_view}
  \end{subfigure}%
  \hspace{2pt}%
  \begin{subfigure}[b]{0.913\dvh}
    \centering
    \raisebox{0.069\dvh}{\fbox{\includegraphics[height=\dimexpr0.913\dvh-0.8pt\relax]{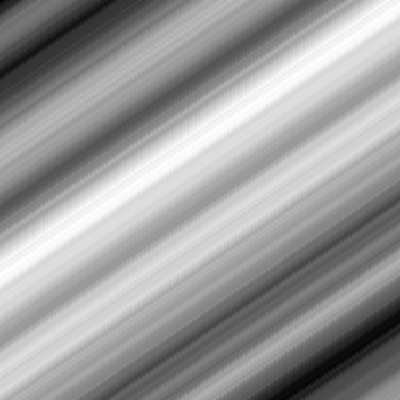}}}
    \caption{Delay-embedding view}
    \label{fig:delay_view}
  \end{subfigure}
  \caption{Dual-view visual representations of a time-series variate. (a) The line plot retains the original measurement frame and axis values. (b) The normalized delay map arranges overlapping windows as columns, so recurring structure appears as diagonal stripes.}
  \label{fig:dual_view_representation}
\end{figure}

Figure~\ref{fig:dual_view_representation} illustrates the two visual representations constructed from the same input time series. The line-plot view (left) preserves the original measurement frame, including absolute amplitudes and axis information. In contrast, the normalized delay-embedding view (right) represents overlapping temporal trajectories after per-series normalization, thereby removing offsets and positive amplitude rescaling while emphasizing temporal structure. The two views are provided to their respective visual encoders as described in Section~\ref{sec:representations}.

\subsection{Delay-Map Construction Details}
\label{app:delay_details}

\paragraph{Hyperparameters.}
Table~\ref{tab:delay_hparams} lists the hyperparameters of the delay map (Eq.~\ref{eq:delay_map}). With these values, the window length $\ell$ equals the series length for $m<48$, is fixed at $\ell_{\min}=48$ for $48\le m<80$, scales as $\lfloor 0.6m\rfloor$ for $80\le m\le 320$, and is capped at $\ell_{\max}=192$ for longer series.

\begin{table}[h]
\centering
\caption{Hyperparameters of the normalized delay-embedding view.}
\label{tab:delay_hparams}
\begin{tabular}{llc}
\toprule
Symbol & Description & Value \\
\midrule
$\rho$ & Window length as a fraction of series length & $0.6$ \\
$\ell_{\min}$ & Minimum window length & $48$ \\
$\ell_{\max}$ & Maximum window length & $192$ \\
$W$ & Number of windows (map width) & $256$ \\
$H$ & Resampled window length (map height) & $256$ \\
\bottomrule
\end{tabular}
\end{table}

\paragraph{Interpolation.}
Interpolation is applied only within each window: the $\ell$ values of $V_c$ are placed on a uniform grid over $[0,1]$ and linearly interpolated at $H$ uniformly spaced points. Window start indices $p_c$ are rounded to integers, so no interpolation is performed along the time axis. Each entry of the map is therefore an interpolated sample of the original series rather than a resized pixel. The resulting $H\times W$ matrix is min--max normalized, replicated across three channels, and normalized with the DINOv3 preprocessing statistics as a floating-point tensor, without quantization or image resizing.

\paragraph{Relation to the standard delay embedding.}
The classical delay embedding~\citep{takens2006detecting, naiman2024utilizing} arranges windows of fixed dimension $n$ as columns with a fixed integer skip $s$ between their start indices, yielding an $n\times\lceil (m-n)/s\rceil$ matrix that is zero-padded to the network input size. Our construction corresponds to $n=\ell$ with a length-dependent fractional skip $(m-\ell)/(W-1)$. Instead of padding, both dimensions are fixed: the number of columns to $W$ and the number of rows to $H$ through resampling. When $m-\ell<W-1$, the skip is below one sample and neighboring columns may repeat. When $\ell=m$, all windows start at $p_c=0$, and the map reduces to $W$ identical copies of the resampled series.

\section{Model and Training Details}
\label{app:model_training}

\subsection{Implementation Details}
\label{app:implementation}

\paragraph{Line-plot tokens.} Each chart is rendered at $768 \times 320$~px and ingested by the Qwen3-VL vision encoder at $736 \times 320$~px, corresponding to $46 \times 20$ patches. After $2\times2$ merging, this yields $N_i^{\mathrm{plot}}=230$ visual tokens per series. The Qwen3-VL tokenizer frames these with two boundary tokens (\texttt{<|vision\_start|>} and \texttt{<|vision\_end|>}) enclosing 230 \texttt{<|image\_pad|>} slots, for an exact budget of 232 tokens per series. The native encoder additionally provides three intermediate feature maps, which Qwen3-VL injects into the language model through DeepStack.

\paragraph{Delay tokens.} For each time-series variate, the delay map $V^{\mathrm{delay}} \in [0,1]^{256\times256}$ is replicated across three channels and normalized using the DINOv3 image preprocessing statistics before being passed to the encoder. We retain the spatial patch tokens and discard the class and register tokens. The resulting $16\times16$ patch grid is merged into an $8\times8$ grid of $2\times2$ groups and projected by the two-layer module $M^{(0)}$, giving $N^{\mathrm{delay}} = 64$ visual tokens per variate; with the two boundary tokens, the exact budget is $66$ tokens per variate.

\paragraph{Multi-level delay features.} The intermediate delay features used for DeepStack fusion are computed as
\begin{equation}
    \mathbf{D}_i = M^{(i)}\big(E^{(k_i)}_{\mathrm{delay}}(V^{\mathrm{delay}})\big), \qquad (k_1,k_2,k_3) = (6,12,18),
\end{equation}
where $E^{(k_i)}_{\mathrm{delay}}$ denotes the output of transformer block $k_i$ and each $M^{(i)}$ applies the same $2\times2$ merge and two-layer projection as $M^{(0)}$, with separate parameters. $\mathbf{D}_1$, $\mathbf{D}_2$, and $\mathbf{D}_3$ are added to the hidden states of the first, second, and third language-model layers, respectively, at the delay-token positions.

\subsection{Training Objective and Hyperparameters}
\label{app:training_details}

For both stages, letting $\theta_s$ denote the parameters optimized at stage $s$, we minimize the autoregressive language-modeling loss
\begin{equation}
\mathcal{L}_{\mathrm{LM}}(\theta_s) = -\sum_{j=1}^{|A|} \log P_{\theta_s}
\left(a_j \mid a_{<j},Q,\mathcal{X} \right) \ ,
\label{eq:training_objective}
\end{equation}
where the loss is computed only over target answer tokens; textual prompts, visual tokens, and padding positions are masked. In Stage~I, $\theta_1$ comprises the DINOv3 LoRA parameters, the delay-token projection $M^{(0)}$, and the three DeepStack projections $\{M^{(i)}\}_{i=1}^{3}$ applied to the features of blocks $(k_1,k_2,k_3)=(6,12,18)$. In Stage~II, $\theta_2$ comprises the LoRA parameters of the language model. In Stage~I, we apply LoRA with rank $r=64$ and scaling parameter $\alpha=128$ to the DINOv3 backbone, while the projections $\{M^{(i)}\}_{i=0}^{3}$ are trained in full. In Stage~II, the Stage~I adapter is merged into DINOv3, and we apply LoRA with rank $r=32$, $\alpha=64$, and dropout $0.05$ to all attention and MLP projections of the language model ($q$, $k$, $v$, $o$, gate, up, and down).

\section{Additional Experimental Results}
\label{app:full_results}

This appendix provides additional results supporting the analyses in the main
paper. We first summarize the representation ablations across benchmarks,
followed by detailed ChatTS results, the complete intervention study, and
experiments with an alternative frequency-domain representation. Together,
these experiments examine whether the gains of the dual-view model extend
beyond TSRBench, whether they can be reproduced by simply duplicating a
representation, and whether the choice of the complementary representation
matters.

\subsection{Cross-Benchmark Representation Ablations}
\label{app:cross_benchmark_ablations}

Table~\ref{tab:cross_benchmark_ablations} extends the representation ablations
beyond TSRBench. We compare the full dual-view model with single-view variants
(\textit{Plot-only} and \textit{Delay-only}) and duplicated-view controls
(\textit{Plot+Plot} and \textit{Delay+Delay}). The duplicated-view controls
keep the full dual-tower architecture, per-tower token budgets, and training
procedure of our model, but feed both towers the same representation; they
therefore test whether the improvement can be reproduced by a second visual
stream alone (Appendix~\ref{app:token_budget}).

The results show that the contribution of the complementary representation
depends on the benchmark. Plot only is already strong on TSExam and ChatTS,
whereas Delay only is substantially weaker on numerical alignment. Combining
the representations provides modest improvements over Plot only on ChatTS
numerical alignment and reasoning, while producing a substantially larger
improvement on the broader TSRBench evaluation. Importantly, neither
duplicated-view control reproduces the overall TSRBench gain of the
complementary pair, despite matching its architecture and visual-token budget.

\begin{table*}[t]
\centering
\small
\setlength{\tabcolsep}{5pt}
\caption{
Cross-benchmark representation ablations. ChatTS-A and ChatTS-B report
categorical (Cat.) and numerical (Num.) alignment accuracy. ChatTS Reason.
reports average accuracy across the four reasoning evaluations. TSRBench
reports overall accuracy. Plot+Plot and Delay+Delay feed the same
representation to both towers while matching the architecture and
visual-token budget of our model.
}
\label{tab:cross_benchmark_ablations}
\begin{tabular}{lccccccc}
\toprule
& \textbf{TSExam} &
\multicolumn{2}{c}{\textbf{ChatTS-A}} &
\multicolumn{2}{c}{\textbf{ChatTS-B}} &
\textbf{ChatTS} &
\textbf{TSRBench} \\
\cmidrule(lr){3-4}
\cmidrule(lr){5-6}
Model & Overall & Cat. & Num. & Cat. & Num. & Reason. & Overall \\
\midrule
Delay only & 0.749 & 0.782 & 0.290 & 0.806 & 0.308 & 0.596 & 0.405 \\
Plot only & \textbf{0.899} & 0.861 & 0.781 & \textbf{0.895} & 0.864 & 0.613 & 0.452 \\
Delay+Delay & 0.790 & 0.858 & 0.424 & 0.848 & 0.348 & 0.604 & 0.426 \\
Plot+Plot & 0.865 & \textbf{0.873} & \textbf{0.805} & 0.877 & 0.859 & 0.586 & 0.448 \\
\midrule
Ours & \textbf{0.899} & 0.864 & \textbf{0.805} & 0.893 & \textbf{0.868} & \textbf{0.615} & \textbf{0.532} \\
\bottomrule
\end{tabular}
\end{table*}

\subsection{TSRBench Category-Level Ablations}
\label{app:tsrbench_category_ablations}

Although Table~\ref{tab:tsrbench_main} reports all 15 individual TSRBench
tasks, Table~\ref{tab:tsrbench_category_ablations} provides the corresponding
category-level aggregation. The dual-view model leaves the already strong
perception performance of Plot only essentially unchanged, while improving
the aggregate results on reasoning, prediction, and decision-making. The
same-view controls again remain close to their corresponding single-view
variants in overall accuracy.

\begin{table}[t]
\centering
\small
\setlength{\tabcolsep}{4pt}
\caption{
TSRBench category-level accuracy (\%) for the representation ablations.
Overall accuracy is computed using the benchmark's complete task-level
evaluation.
}
\label{tab:tsrbench_category_ablations}
\begin{tabular}{lccccc}
\toprule
Model & Perception & Reasoning & Prediction & Decision & Overall \\
\midrule
Delay only & 63.7 & 36.5 & 39.4 & 27.6 & 40.5 \\
Plot only & \textbf{87.4} & 35.8 & 38.0 & 36.4 & 45.2 \\
Delay+Delay & 71.6 & 37.0 & 41.7 & 27.4 & 42.6 \\
Plot+Plot & 85.3 & 36.3 & 40.3 & 30.7 & 44.8 \\
\midrule
Ours & 86.4 & \textbf{40.1} & \textbf{61.7} & \textbf{37.6} & \textbf{53.2} \\

\bottomrule
\end{tabular}
\end{table}

\subsection{Detailed ChatTS Results}
\label{app:chatts_detailed}

We further examine the representation ablations on the ChatTS reasoning
evaluation. Table~\ref{tab:chatts_reasoning_ablations} reports the four
reasoning dimensions individually rather than only their average. The results
show that no representation uniformly dominates every reasoning dimension.
The full model achieves the highest average accuracy, while individual
single-view variants remain stronger on some dimensions. This variation is
consistent with the central premise that the representations provide
different inductive biases rather than one view uniformly subsuming the
other.

\begin{table}[t]
\centering
\small
\setlength{\tabcolsep}{5pt}
\caption{
ChatTS reasoning accuracy for the single-view and dual-view models on
inductive, deductive, and causal reasoning and two-series multiple-choice
questions (MCQ2). Inductive reasoning is scored with the RAGAS pipeline;
Average reports the mean over the four tasks.
}
\label{tab:chatts_reasoning_ablations}
\begin{tabular}{lccccc}
\toprule
Model & Induct. & Deduct. & Causal & MCQ2 & Average \\
\midrule
Delay only & 0.403 & 0.605 & 0.857 & 0.520 & 0.596 \\
Plot only & 0.441 & 0.548 & 0.851 & 0.610 & 0.613 \\
Ours & 0.479 & 0.535 & 0.826 & 0.620 & 0.615 \\
\bottomrule
\end{tabular}
\end{table}

The alignment results in Table~\ref{tab:cross_benchmark_ablations} provide a
complementary picture. Delay only performs considerably worse on numerical
alignment than on categorical alignment, whereas Plot only retains strong
performance on both. The dual representation preserves the strong categorical
performance of Plot only while improving numerical accuracy from 0.781 to
0.805 on ChatTS-A and from 0.864 to 0.868 on ChatTS-B.

\subsection{Complete Intervention Results}
\label{app:interventions}

Table~\ref{tab:tsrb_interventions_effective} expands the intervention analysis from
Section~\ref{sec:robustness_analysis} by reporting both absolute accuracy and
the percentage-point change relative to clean performance. Unlike the main analysis, which is restricted to the 2{,}020 scale- and sampling-invariant items, this table covers all 15 TSRBench tasks. We additionally report two temporal-symmetry transformations, temporal reversal and sign inversion. Unlike the other interventions, these can legitimately change the gold answer (e.g., reversing an increasing trend), so they illustrate that invariance is not always desirable: a model that is insensitive to these transformations may be ignoring information the task requires.

The full results make the distinct sensitivity profiles particularly clear.
Delay only changes by at most 0.8 percentage points under the five affine
interventions, including $z$-score normalization, which is consistent with
the invariance induced by its per-series normalization. In
contrast, Plot only and our model remain sensitive to transformations of the
measurement frame. Delay only and Plot only also exhibit the smallest average degradation
under noise and smoothing. Our model is not uniformly the most invariant
representation, but retains substantially higher absolute accuracy across all
reported interventions. Under changes in sampling horizon, its category-average
accuracy is virtually unchanged from clean performance.

\begin{table*}[tbp]
\centering
\small
\setlength{\tabcolsep}{3.8pt}
\caption{
Evaluation-time affine and dynamical interventions across the full TSRBench benchmark (all 15 tasks, including forecasting). For each model, we report clean accuracy and the percentage-point change ($\Delta$) relative to clean
performance. Daggers ($\dagger$) mark non-invariance interventions for which
gold targets legitimately change. Bold values indicate the smallest absolute
shift ($|\Delta|$) in each row.
}
\label{tab:tsrb_interventions_effective}
\begin{tabular}{ll cc cc cc}
\toprule
& & \multicolumn{2}{c}{\textbf{Plot only}} & \multicolumn{2}{c}{\textbf{Ours}} & \multicolumn{2}{c}{\textbf{Delay only}} \\
\cmidrule(lr){3-4}
\cmidrule(lr){5-6}
\cmidrule(lr){7-8}
Intervention Type & Transformation & Acc. & $\Delta$ vs. Clean & Acc. & $\Delta$ vs. Clean & Acc. & $\Delta$ vs. Clean \\
\midrule
Baseline & Clean & 0.452 & --- & 0.532 & --- & 0.405 & --- \\
\midrule
Affine Invariance & Scale $\times 0.1$ & 0.467 & $+1.5$ & 0.496 & $-3.6$ & 0.407 & $\mathbf{+0.2}$ \\
& Scale $\times 10$ & 0.439 & $-1.4$ & 0.522 & $-1.0$ & 0.411 & $\mathbf{+0.6}$ \\
& Offset $+5\sigma$ & 0.431 & $-2.1$ & 0.511 & $-2.2$ & 0.397 & $\mathbf{-0.8}$ \\
& Offset $-5\sigma$ & 0.430 & $-2.2$ & 0.520 & $-1.3$ & 0.400 & $\mathbf{-0.6}$ \\
& $z$-score & 0.418 & $-3.4$ & 0.499 & $-3.3$ & 0.397 & $\mathbf{-0.8}$ \\
\cmidrule(lr){2-8}
& \textit{Category Average} & 0.437 & $-1.5$ & 0.509 & $-2.3$ & 0.403 & $\mathbf{-0.3}$ \\
\midrule
Noise \& Smoothing & Additive Noise ($20\text{ dB}$) & 0.440 & $\mathbf{-1.2}$ & 0.520 & $-1.2$ & 0.390 & $-1.6$ \\
& Additive Noise ($10\text{ dB}$) & 0.416 & $-3.7$ & 0.496 & $-3.6$ & 0.371 & $\mathbf{-3.4}$ \\
& Additive Noise ($0\text{ dB}$) & 0.407 & $\mathbf{-4.6}$ & 0.481 & $-5.1$ & 0.359 & $-4.7$ \\
& Smoothing ($w=5$) & 0.434 & $-1.8$ & 0.494 & $-3.9$ & 0.389 & $\mathbf{-1.6}$ \\
\cmidrule(lr){2-8}
& \textit{Category Average} & 0.424 & $\mathbf{-2.8}$ & 0.498 & $-3.4$ & 0.377 & $-2.8$ \\
\midrule
Temporal Symmetry & Temporal Reversal$^\dagger$ & 0.426 & $-2.6$ & 0.500 & $-3.2$ & 0.398 & $\mathbf{-0.8}$ \\
& Sign Inversion$^\dagger$ & 0.427 & $-2.5$ & 0.510 & $-2.2$ & 0.404 & $\mathbf{-0.1}$ \\
\cmidrule(lr){2-8}
& \textit{Category Average} & 0.427 & $-2.6$ & 0.505 & $-2.7$ & 0.401 & $\mathbf{-0.4}$ \\
\midrule
Sampling Horizon & Resample ($256$ pts) & 0.442 & $-1.1$ & 0.524 & $\mathbf{-0.8}$ & 0.423 & $+1.7$ \\
& Resample ($2048$ pts) & 0.447 & $-0.6$ & 0.534 & $\mathbf{+0.1}$ & 0.401 & $-0.4$ \\
\cmidrule(lr){2-8}
& \textit{Category Average} & 0.444 & $-0.8$ & 0.529 & $\mathbf{-0.3}$ & 0.412 & $+0.7$ \\
\bottomrule
\end{tabular}
\end{table*}

\subsection{Alternative Secondary Representations}
\label{app:alternative_representation}

The main model uses a normalized delay representation as the complementary
view. To test whether the observed gains follow simply from introducing any
second representation, we additionally replace the delay view with a
short-time Fourier transform (STFT) representation while retaining the
line-plot stream. We evaluate this alternative both in a controlled setting
and after full training.

\paragraph{Controlled representation comparison.}
We first compare delay and STFT representations under the same controlled
TSExam setting. Table~\ref{tab:alternative_rep_ts_exam} reports overall
accuracy for the secondary representation alone and when combined with the
line-plot representation. Both raw and normalized STFT variants are included.
The results show that the three secondary representations are similar when
combined with the line plot on this evaluation, indicating that TSExam alone
does not strongly distinguish among them.

\begin{table}[t]
\centering
\small
\setlength{\tabcolsep}{7pt}
\caption{
Controlled TSExam comparison of alternative secondary representations.
``Only'' evaluates the secondary representation without the line-plot stream,
whereas ``+ Plot'' combines it with the line plot.
}
\label{tab:alternative_rep_ts_exam}
\begin{tabular}{lcc}
\toprule
Secondary Representation & Only & + Plot \\
\midrule
STFT ($z$-normalized) & 0.739 & 0.889 \\
STFT (raw) & 0.745 & 0.894 \\
Delay & 0.737 & 0.893 \\
\bottomrule
\end{tabular}
\end{table}

\paragraph{Robustness to evaluation-time transformations.}
We further apply deterministic transformations to the TSExam inputs of the
single-representation (``Only'') models from
Table~\ref{tab:alternative_rep_ts_exam}. As shown in
Table~\ref{tab:alternative_rep_interventions}, the representations exhibit
different robustness profiles. The delay embedding and the $z$-normalized STFT
are nearly invariant to scaling and offsets, whereas the raw STFT degrades
under mean shifts. The raw STFT is the most robust to additive noise, while
the delay embedding is the most robust to smoothing and to resampling at a
longer horizon, but the most sensitive to temporal reversal and sign
inversion.

\begin{table*}[tbp]
\centering
\small
\setlength{\tabcolsep}{3.8pt}
\caption{
Evaluation-time affine and dynamical interventions on TSExam ($N=746$) for the
single-representation models. For each model, we report clean accuracy and the
percentage-point change ($\Delta$) relative to clean performance. Daggers
($\dagger$) mark non-invariance interventions for which gold targets
legitimately change. Bold values indicate the smallest absolute shift
($|\Delta|$) in each row.
}
\label{tab:alternative_rep_interventions}
\begin{tabular}{ll cc cc cc}
\toprule
& & \multicolumn{2}{c}{\textbf{STFT ($z$-normalized)}} & \multicolumn{2}{c}{\textbf{STFT (raw)}} & \multicolumn{2}{c}{\textbf{Delay}} \\
\cmidrule(lr){3-4}
\cmidrule(lr){5-6}
\cmidrule(lr){7-8}
Intervention Type & Transformation & Acc. & $\Delta$ vs. Clean & Acc. & $\Delta$ vs. Clean & Acc. & $\Delta$ vs. Clean \\
\midrule
Baseline & Clean & 0.739 & --- & 0.745 & --- & 0.737 & --- \\
\midrule
Affine Invariance & Scale $\times 0.1$ & 0.725 & $-1.3$ & 0.727 & $-1.9$ & 0.733 & $\mathbf{-0.4}$ \\
& Scale $\times 10$ & 0.728 & $-1.1$ & 0.727 & $-1.9$ & 0.731 & $\mathbf{-0.7}$ \\
& Offset $+5\sigma$ & 0.728 & $-1.1$ & 0.634 & $-11.1$ & 0.732 & $\mathbf{-0.5}$ \\
& Offset $-5\sigma$ & 0.729 & $-0.9$ & 0.649 & $-9.7$ & 0.731 & $\mathbf{-0.7}$ \\
& $z$-score & 0.724 & $-1.5$ & 0.698 & $-4.7$ & 0.732 & $\mathbf{-0.5}$ \\
\cmidrule(lr){2-8}
& \textit{Category Average} & 0.727 & $-1.2$ & 0.687 & $-5.9$ & 0.732 & $\mathbf{-0.6}$ \\
\midrule
Noise \& Smoothing & Additive Noise ($20\text{ dB}$) & 0.665 & $-7.4$ & 0.736 & $\mathbf{-0.9}$ & 0.708 & $-2.9$ \\
& Additive Noise ($10\text{ dB}$) & 0.538 & $-20.1$ & 0.639 & $\mathbf{-10.6}$ & 0.572 & $-16.5$ \\
& Additive Noise ($0\text{ dB}$) & 0.421 & $-31.8$ & 0.539 & $\mathbf{-20.6}$ & 0.468 & $-26.9$ \\
& Smoothing ($w=5$) & 0.682 & $-5.6$ & 0.653 & $-9.2$ & 0.728 & $\mathbf{-0.9}$ \\
\cmidrule(lr){2-8}
& \textit{Category Average} & 0.577 & $-16.2$ & 0.642 & $\mathbf{-10.3}$ & 0.619 & $-11.8$ \\
\midrule
Temporal Symmetry & Temporal Reversal$^\dagger$ & 0.655 & $-8.3$ & 0.676 & $\mathbf{-7.0}$ & 0.607 & $-13.0$ \\
& Sign Inversion$^\dagger$ & 0.693 & $-4.6$ & 0.744 & $\mathbf{-0.1}$ & 0.631 & $-10.6$ \\
\cmidrule(lr){2-8}
& \textit{Category Average} & 0.674 & $-6.5$ & 0.710 & $\mathbf{-3.6}$ & 0.619 & $-11.8$ \\
\midrule
Sampling Horizon & Resample ($256$ pts) & 0.768 & $+2.9$ & 0.732 & $\mathbf{-1.3}$ & 0.776 & $+3.9$ \\
& Resample ($2048$ pts) & 0.658 & $-8.0$ & 0.653 & $-9.2$ & 0.718 & $\mathbf{-1.9}$ \\
\cmidrule(lr){2-8}
& \textit{Category Average} & 0.713 & $-2.6$ & 0.693 & $-5.3$ & 0.747 & $\mathbf{+1.0}$ \\
\bottomrule
\end{tabular}
\end{table*}

\paragraph{Full-training comparison.}
We next train the STFT-based alternative under the full training setup and
compare it with the proposed delay-based model. As shown in
Table~\ref{tab:stft_full_training}, neither representation uniformly dominates
across all evaluations. STFT is slightly stronger on ChatTS numerical
alignment and the ChatTS reasoning average, whereas the delay-based model is
slightly stronger on TSExam and substantially stronger on the broader
TSRBench evaluation. On ChatTS categorical alignment, the two models are
effectively tied. These results suggest that the benefit of complementary
views is not tied to uniform superiority of the delay representation on every
task; rather, the choice of secondary representation changes the inductive
bias available to the language model.

\begin{table}[t]
\centering
\small
\setlength{\tabcolsep}{5pt}
\caption{
Full-training comparison between the proposed delay-based dual-view model and
an alternative model using STFT as the secondary representation. ChatTS-A and
ChatTS-B report categorical (Cat.) and numerical (Num.) alignment accuracy;
Reason. reports the ChatTS reasoning average.
}
\label{tab:stft_full_training}
\begin{tabular}{lccccccc}
\toprule
 & & \multicolumn{2}{c}{ChatTS-A} & \multicolumn{2}{c}{ChatTS-B} & & \\
\cmidrule(lr){3-4}\cmidrule(lr){5-6}
Model & TSExam & Cat. & Num. & Cat. & Num. & Reason. & TSRBench \\
\midrule
Plot + STFT & 0.890 & 0.859 & 0.809 & 0.894 & 0.871 & 0.628 & 0.467 \\
Plot + Delay (Ours) & 0.899 & 0.864 & 0.805 & 0.893 & 0.868 & 0.615 & 0.532 \\
\bottomrule
\end{tabular}
\end{table}

\subsection{Visual-Token Budget and Same-View Controls}
\label{app:token_budget}

For each input series, the line-plot representation is encoded by the native Qwen3-VL vision encoder into 230 visual tokens, while the delay representation contributes 64 visual tokens, for a total of 294 visual tokens per series in the full model. The \textit{Plot+Plot} and \textit{Delay+Delay} controls preserve this budget and the complete dual-tower architecture, changing only the input to one tower. In \textit{Plot+Plot}, the DINOv3 tower receives the line plot, rasterized at $256\times256$~px, in place of the delay map, yielding 64 tokens. In \textit{Delay+Delay}, the Qwen3-VL tower receives the delay map rendered at approximately $480\times480$~px, yielding approximately 230 tokens, with the embedding window scaled so that its geometry matches the $256$-px DINOv3 input. Each control is trained from its own Stage~I alignment followed by Stage~II instruction tuning, using the same data and hyperparameters as our model. The controls are therefore matched to the full model in architecture, trainable parameters, visual-token budget, and training procedure, and differ only in whether the two towers receive distinct or identical views. They are not matched to the single-view variants, which use one tower and fewer visual tokens. Notably, Delay+Delay allocates substantially more tokens to the delay representation than our model does, and Plot+Plot allocates more tokens to the line plot, yet neither reproduces the gain of the complementary pair, indicating that the improvement cannot be attributed to additional capacity devoted to either view alone.

\subsection{Complete Benchmark Results}
\label{app:full_tables}
This section reports the complete per-task results underlying the summaries in the main text. Table~\ref{tab:tsrbench_app} gives TSRBench accuracy for every sub-task, grouped by input modality (text, vision, both, and dedicated TSLMs). Table~\ref{tab:chatts_app} reports the ChatTS alignment benchmark on univariate and multivariate series, and Table~\ref{tab:chatts_reasoning_app} reports the ChatTS reasoning evaluation. Table~\ref{tab:tsexam_app} breaks down TSExam accuracy by question category. Unless otherwise noted, baseline results are taken from the original benchmark papers. We evaluated GPT-5 (\texttt{gpt-5-2025-08-07}) on TSExam and ChatTS ourselves with text (T), vision (V), and combined (T+V) inputs, using the default sampling parameters and the default (medium) reasoning effort. For TSExam, we use the same minimal prompt as in all of our evaluations: the question and its options, with an instruction to reply with the option letter alone, and without the optional hints, concept definitions, example series, worked example, or step-by-step reasoning instruction. Responses are scored by the selected option letter. On TSRBench, we re-evaluated the Qwen vision-language models (Qwen2.5-VL-7B, Qwen2.5-VL-72B, Qwen3-VL-8B, and Qwen3-VL-32B) ourselves, rendering each time series with the same line-plot pipeline used to evaluate our model, so that the comparison with our Qwen3-VL-based model is not confounded by differences in plot rendering. All other TSRBench baselines, including GPT-5, are taken from the benchmark.

\begin{table*}[tbp]\centering\scriptsize
\setlength{\tabcolsep}{2pt}
\caption{TSRBench accuracy (\%) by input modality. Bold columns (Perc., Reas., Pred., Dec.) are item-weighted category averages over their sub-tasks. Gray rows denote reference models (proprietary or open-weight models larger than 72B parameters).}
\label{tab:tsrbench_app}
\begin{tabular}{l ccccc cccccccc ccc ccc c}
\toprule
 & \multicolumn{5}{c}{Perception} & \multicolumn{8}{c}{Reasoning} & \multicolumn{3}{c}{Prediction} & \multicolumn{3}{c}{Decision} & \\
\cmidrule(lr){2-6}\cmidrule(lr){7-14}\cmidrule(lr){15-17}\cmidrule(lr){18-20}
Model & \textbf{Perc.} & PR & NU & AD & CA & \textbf{Reas.} & ER & CD & AR & TR & NR & DR & IR & \textbf{Pred.} & TSF & EP & \textbf{Dec.} & QualDM & QuantDM & \textbf{Overall} \\
\midrule
\multicolumn{21}{c}{\textit{Textual Time Series as Input}} \\
\midrule
\rowcolor{gray!15} DeepSeek-V3.2 & 63.9 & 67.7 & 56.3 & 57.4 & 64.6 & 35.6 & 32.0 & 35.7 & 70.7 & 19.4 & 27.3 & 47.6 & 24.0 & 33.4 & 26.5 & 47.2 & 30.8 & 33.1 & 28.3 & 39.1 \\
\rowcolor{gray!15} GPT-OSS-120B & 64.6 & 66.8 & 56.3 & 59.8 & 69.0 & 38.5 & 39.1 & 26.3 & 58.7 & 31.3 & 48.5 & 28.0 & 40.0 & 40.6 & 31.1 & 59.7 & 32.4 & 33.7 & 31.0 & 42.0 \\
\rowcolor{gray!15} Qwen3-235B-A22B & 63.8 & 66.0 & 56.3 & 59.7 & 67.3 & 40.9 & 22.6 & 34.7 & 86.7 & 28.1 & 44.8 & 49.2 & 39.0 & 35.6 & 29.0 & 48.9 & 32.5 & 34.8 & 30.0 & 42.2 \\
\rowcolor{gray!15} GPT-5-mini & 68.7 & 72.2 & 59.8 & 63.6 & 69.9 & 47.5 & 45.1 & 27.7 & 62.7 & 39.4 & 65.2 & 41.3 & 50.0 & 38.3 & 23.5 & 67.8 & 33.0 & 35.5 & 30.3 & 46.6 \\
\rowcolor{gray!15} o4-mini & 69.7 & 73.1 & 65.5 & 61.2 & 71.7 & 47.7 & 42.6 & 39.0 & 58.0 & 34.4 & 64.7 & 42.0 & 43.0 & 46.5 & 33.1 & 73.3 & 33.0 & 30.4 & 36.0 & 47.7 \\
\rowcolor{gray!15} GPT-5 & 70.3 & 75.7 & 62.1 & 61.2 & 69.0 & 60.1 & 42.3 & 55.3 & 85.3 & 68.8 & 72.2 & 50.0 & 62.0 & 51.8 & 37.8 & 79.7 & 31.9 & 31.9 & 32.0 & 55.5 \\
Qwen2.5-3B & 45.4 & 46.4 & 51.7 & 33.3 & 51.3 & 27.3 & 20.0 & 25.0 & 38.0 & 21.2 & 34.8 & 29.2 & 19.0 & 40.4 & 31.4 & 58.3 & 23.3 & 22.7 & 24.0 & 33.2 \\
Qwen2.5-7B & 49.6 & 50.7 & 50.6 & 41.1 & 54.9 & 28.3 & 15.1 & 32.3 & 46.0 & 28.1 & 33.2 & 24.0 & 28.0 & 34.8 & 33.5 & 37.5 & 28.5 & 31.0 & 25.7 & 33.7 \\
Qwen3-1.7B & 47.0 & 45.8 & 59.8 & 38.0 & 51.3 & 27.7 & 18.3 & 30.1 & 48.7 & 30.0 & 27.0 & 28.4 & 20.0 & 48.7 & 39.2 & 67.8 & 28.2 & 31.3 & 24.7 & 36.8 \\
InternLM3-8B & 47.7 & 51.2 & 42.5 & 35.7 & 54.0 & 28.6 & 23.4 & 24.7 & 48.0 & 23.8 & 30.0 & 28.7 & 31.0 & 46.5 & 36.6 & 66.4 & 30.4 & 33.7 & 26.7 & 37.0 \\
Gemma3-12B-it & 54.3 & 51.5 & 59.8 & 48.9 & 65.5 & 30.6 & 21.1 & 34.7 & 42.0 & 23.7 & 33.2 & 28.0 & 41.0 & 37.7 & 39.4 & 34.2 & 32.9 & 34.3 & 31.3 & 37.2 \\
GPT-OSS-20B & 61.7 & 64.2 & 58.6 & 50.4 & 69.0 & 33.6 & 31.4 & 32.0 & 48.7 & 17.5 & 43.3 & 20.7 & 43.0 & 30.7 & 27.4 & 37.2 & 33.9 & 34.9 & 32.7 & 38.1 \\
Qwen3-8B & 52.6 & 51.8 & 56.3 & 48.8 & 56.6 & 29.1 & 14.0 & 29.3 & 57.3 & 23.1 & 36.5 & 23.6 & 33.0 & 47.4 & 46.7 & 48.9 & 32.0 & 34.9 & 28.7 & 38.3 \\
Qwen3-32B & 55.0 & 54.7 & 52.9 & 50.4 & 62.8 & 35.2 & 22.6 & 35.9 & 66.0 & 27.5 & 37.5 & 34.4 & 36.0 & 36.1 & 31.1 & 46.1 & 33.4 & 34.9 & 31.7 & 38.6 \\
Gemma3-27B-it & 57.0 & 56.1 & 59.8 & 48.1 & 68.1 & 33.7 & 22.3 & 37.0 & 66.7 & 21.9 & 36.0 & 29.3 & 35.0 & 38.3 & 36.1 & 42.8 & 31.8 & 33.1 & 30.3 & 38.6 \\
Qwen2.5-72B & 55.6 & 55.3 & 55.2 & 47.3 & 66.4 & 39.7 & 20.9 & 58.0 & 62.0 & 33.8 & 38.2 & 36.8 & 40.0 & 43.9 & 30.7 & 70.3 & 32.3 & 34.0 & 30.3 & 42.4 \\
\midrule
\multicolumn{21}{c}{\textit{Visual Time Series as Input}} \\
\midrule
\rowcolor{gray!15} Qwen3-VL-235B-A22B & 65.9 & 65.8 & 65.5 & 61.2 & 71.7 & 44.4 & 39.7 & 47.7 & 84.7 & 25.0 & 43.0 & 42.8 & 31.0 & 25.7 & 25.3 & 26.4 & 30.7 & 32.5 & 28.7 & 41.0 \\
\rowcolor{gray!15} Llama-4-Scout-17B-16E & 42.3 & 41.5 & 44.8 & 38.8 & 46.9 & 39.1 & 25.1 & 53.0 & 75.3 & 28.1 & 41.5 & 30.0 & 23.0 & 52.5 & 39.7 & 78.0 & 33.5 & 33.7 & 33.3 & 42.3 \\
\rowcolor{gray!15} GPT-5-mini & 74.1 & 78.7 & 69.0 & 69.0 & 69.0 & 45.2 & 36.0 & 41.0 & 58.7 & 41.9 & 54.5 & 39.6 & 51.0 & 40.4 & 27.5 & 66.1 & 27.1 & 26.0 & 28.3 & 46.0 \\
\rowcolor{gray!15} o4-mini & 74.2 & 77.4 & 72.4 & 66.7 & 73.5 & 40.9 & 29.7 & 30.7 & 55.3 & 39.4 & 56.2 & 32.8 & 51.0 & 44.1 & 29.4 & 73.6 & 35.8 & 29.9 & 42.3 & 46.6 \\
\rowcolor{gray!15} GPT-5 & 78.0 & 83.6 & 72.4 & 69.8 & 73.5 & 52.4 & 34.6 & 37.0 & 81.3 & 68.1 & 64.0 & 48.0 & 57.0 & 49.8 & 38.8 & 71.9 & 28.4 & 26.9 & 30.0 & 52.4 \\
Qwen2.5-VL-3B & 46.0 & 44.2 & 50.6 & 41.9 & 53.1 & 27.6 & 23.4 & 24.0 & 42.0 & 25.6 & 30.2 & 27.6 & 25.0 & 30.9 & 24.2 & 44.2 & 25.8 & 31.0 & 20.0 & 31.3 \\
Phi4-Multimodal-8B & 49.0 & 52.3 & 46.0 & 41.9 & 48.7 & 26.0 & 24.6 & 22.3 & 28.7 & 23.1 & 30.5 & 25.2 & 26.0 & 30.0 & 32.1 & 25.8 & 32.1 & 34.9 & 29.0 & 31.9 \\
InternVL3.5-1B & 45.9 & 47.2 & 49.4 & 38.8 & 46.9 & 27.8 & 25.7 & 24.3 & 46.7 & 24.4 & 29.2 & 24.0 & 27.0 & 38.6 & 34.7 & 46.4 & 28.2 & 34.3 & 21.3 & 33.8 \\
Qwen2.5-VL-7B & 54.7 & 55.0 & 52.9 & 52.7 & 57.5 & 28.0 & 15.1 & 28.0 & 54.0 & 33.1 & 28.0 & 26.8 & 28.0 & 44.1 & 38.6 & 55.0 & 21.9 & 36.4 & 5.7 & 35.8 \\
MiniCPM-V-4.5-8B & 61.3 & 63.6 & 56.3 & 56.6 & 62.8 & 26.5 & 22.3 & 24.3 & 54.7 & 20.6 & 26.5 & 22.8 & 24.0 & 39.0 & 44.6 & 27.8 & 27.9 & 26.3 & 29.7 & 35.9 \\
MiMo-VL-7B-RL & 59.1 & 58.2 & 64.4 & 52.7 & 65.5 & 33.1 & 23.7 & 34.7 & 65.3 & 30.6 & 33.0 & 25.6 & 36.0 & 33.2 & 35.0 & 29.7 & 30.9 & 29.0 & 33.0 & 37.2 \\
Qwen2.5-VL-72B & 63.9 & 63.1 & 55.2 & 59.7 & 77.9 & 37.1 & 27.4 & 52.3 & 58.0 & 33.1 & 39.5 & 24.0 & 23.0 & 35.3 & 16.7 & 72.5 & 29.0 & 29.3 & 28.7 & 39.9 \\
InternVL3.5-38B & 60.3 & 58.8 & 60.9 & 55.8 & 69.9 & 38.1 & 30.9 & 43.0 & 52.7 & 31.2 & 43.8 & 30.4 & 34.0 & 32.3 & 32.2 & 32.5 & 32.3 & 35.2 & 29.0 & 39.4 \\
InternVL3.5-8B & 59.3 & 60.9 & 55.2 & 52.7 & 64.6 & 35.6 & 25.7 & 38.3 & 60.0 & 27.5 & 40.5 & 31.6 & 29.0 & 39.4 & 38.2 & 41.9 & 28.2 & 33.4 & 22.3 & 39.5 \\
Qwen3-VL-8B & 58.3 & 54.7 & 57.5 & 62.8 & 65.5 & 29.5 & 24.3 & 31.7 & 52.0 & 23.1 & 33.2 & 19.2 & 29.0 & 44.6 & 51.7 & 30.6 & 32.8 & 36.1 & 29.0 & 38.9 \\
Qwen3-VL-32B & 67.3 & 67.4 & 63.2 & 63.6 & 74.3 & 40.4 & 27.7 & 62.3 & 60.0 & 29.4 & 44.8 & 26.4 & 24.0 & 46.2 & 42.6 & 53.3 & 30.9 & 32.8 & 28.7 & 45.0 \\
\midrule
\multicolumn{21}{c}{\textit{Both Visual \& Textual Time Series as Input}} \\
\midrule
\rowcolor{gray!15} Claude-4.5-Haiku & 53.4 & 57.1 & 41.4 & 48.8 & 55.8 & 39.6 & 24.0 & 19.7 & 78.0 & 32.5 & 50.0 & 45.2 & 52.0 & 30.9 & 22.2 & 48.3 & 22.8 & 18.1 & 28.0 & 37.1 \\
\rowcolor{gray!15} Gemini-2.5-Flash & 74.6 & 75.2 & 73.6 & 69.0 & 79.6 & 40.9 & 36.0 & 31.0 & 86.7 & 35.6 & 39.8 & 36.8 & 43.0 & 40.0 & 26.7 & 66.7 & 41.6 & 34.3 & 49.7 & 46.5 \\
\rowcolor{gray!15} GPT-5-mini & 73.1 & 78.4 & 65.5 & 63.6 & 72.6 & 47.1 & 38.9 & 36.3 & 76.0 & 33.1 & 66.2 & 30.0 & 54.0 & 38.1 & 24.4 & 65.6 & 32.1 & 34.3 & 29.7 & 46.9 \\
\rowcolor{gray!15} o4-mini & 74.9 & 79.0 & 70.1 & 65.1 & 76.1 & 45.4 & 37.4 & 38.0 & 64.0 & 35.0 & 66.8 & 24.8 & 51.0 & 44.3 & 29.3 & 74.4 & 32.8 & 29.9 & 36.0 & 48.2 \\
\rowcolor{gray!15} o4-mini-high & 76.7 & 82.5 & 71.3 & 64.3 & 76.1 & 56.7 & 44.0 & 66.3 & 76.7 & 55.0 & 70.8 & 30.8 & 53.0 & 46.4 & 33.6 & 71.9 & 24.9 & 25.4 & 24.3 & 52.5 \\
\rowcolor{gray!15} GPT-5-mini-high & 75.9 & 80.3 & 75.9 & 63.6 & 75.2 & 60.5 & 47.4 & 69.0 & 82.0 & 56.2 & 71.8 & 42.0 & 57.0 & 44.6 & 32.5 & 68.9 & 28.8 & 32.8 & 24.3 & 54.1 \\
\rowcolor{gray!15} GPT-5 & 76.7 & 84.9 & 72.4 & 61.2 & 70.8 & 58.1 & 38.9 & 50.3 & 93.3 & 68.8 & 72.0 & 44.0 & 58.0 & 51.2 & 37.8 & 78.1 & 33.0 & 31.9 & 34.3 & 55.6 \\
\midrule
\multicolumn{21}{c}{\textit{Time Series Language Models (TSLM)}} \\
\midrule
OpenTSLM-tsqa-flamingo-3B & 39.5 & 41.2 & 41.4 & 34.1 & 38.9 & 26.3 & 22.9 & 30.3 & 36.0 & 21.2 & 22.5 & 28.8 & 29.0 & 33.9 & 28.1 & 45.6 & 27.4 & 25.4 & 29.7 & 30.7 \\
OpenTSLM-tsqa-sp-3B & 39.7 & 39.9 & 40.2 & 35.7 & 43.4 & 26.8 & 20.3 & 32.3 & 33.3 & 28.1 & 24.5 & 26.0 & 32.0 & 33.5 & 32.5 & 35.6 & 28.5 & 28.4 & 28.7 & 31.0 \\
ChatTS-14B & 50.7 & 50.7 & 50.6 & 46.5 & 55.8 & 29.3 & 21.7 & 34.3 & 51.3 & 24.4 & 30.5 & 23.6 & 25.0 & 30.2 & 19.2 & 52.2 & 28.5 & 29.3 & 27.7 & 33.5 \\
TS-Reasoner-7B & 52.1 & 53.1 & 56.3 & 48.1 & 50.4 & 31.3 & 28.9 & 24.3 & 57.3 & 26.9 & 37.2 & 23.6 & 24.0 & 39.5 & 31.7 & 55.0 & 27.6 & 32.8 & 21.7 & 36.4 \\
TimeOmni-1-7B & 54.7 & 55.0 & 59.8 & 41.9 & 64.6 & 31.0 & 28.0 & 35.3 & 46.7 & 24.4 & 31.5 & 22.8 & 34.0 & 36.5 & 30.0 & 49.4 & 32.3 & 34.0 & 30.3 & 36.7 \\
\textbf{Ours} & 86.4 & 85.4 & 94.3 & 78.3 & 92.9 & 40.1 & 43.1 & 51.7 & 56.0 & 28.7 & 35.5 & 29.2 & 34.0 & 61.7 & 50.4 & 84.2 & 37.6 & 32.8 & 43.0 & 53.2 \\
\bottomrule
\end{tabular}
\end{table*}

\begin{table*}[tbp]\centering\small
\setlength{\tabcolsep}{3.2pt}
\caption{ChatTS alignment benchmark on univariate (Dataset A) and multivariate (Dataset B) series. Gray rows denote reference models. Corr.\ and Clus.\ are the correlation and clustering sub-tasks (categorical only). Bold values mark the best result per column within each dataset, excluding reference models.}
\label{tab:chatts_app}
\begin{tabular}{l l cc cc cc cc c c cc}
\toprule
Dataset & Model & \multicolumn{2}{c}{Trend} & \multicolumn{2}{c}{Season} & \multicolumn{2}{c}{Noise} & \multicolumn{2}{c}{Local} & Corr. & Clus. & \multicolumn{2}{c}{Overall} \\
\cmidrule(lr){3-4}\cmidrule(lr){5-6}\cmidrule(lr){7-8}\cmidrule(lr){9-10}\cmidrule(lr){11-11}\cmidrule(lr){12-12}\cmidrule(lr){13-14}
 & & Cate. & Num. & Cate. & Num. & Cate. & Num. & Cate. & Num. & Cate. & Cate. & Cate. & Num. \\
\midrule
\rowcolor{gray!15} A & GPT-4o (T) & 0.585 & 0.882 & 0.811 & 0.768 & 0.905 & 0.153 & 0.379 & 0.256 & 0.476 & 0.333 & 0.542 & 0.371 \\
\rowcolor{gray!15}   & GPT-4o (V) & 0.659 & 0.613 & 0.811 & 0.559 & 0.810 & 0.248 & 0.537 & 0.414 & 0.476 & 0.480 & 0.609 & 0.436 \\
\rowcolor{gray!15}   & GPT-5 (T) & 0.512 & 0.887 & 0.973 & 0.000 & 0.905 & 0.350 & 0.905 & 0.741 & 0.881 & 0.485 & 0.797 & 0.724 \\
\rowcolor{gray!15}   & GPT-5 (V) & 0.732 & 0.646 & 0.973 & 0.000 & 0.762 & 0.394 & 0.874 & 0.725 & 0.857 & 0.600 & 0.810 & 0.670 \\
\rowcolor{gray!15}   & GPT-5 (T+V) & 0.585 & 0.905 & 0.973 & 0.000 & 0.905 & 0.423 & 0.863 & 0.735 & 0.857 & 0.558 & 0.801 & 0.731 \\
  & Qwen3-VL-8B (V) & 0.098 & 0.011 & 0.108 & 0.000 & 0.167 & 0.051 & 0.095 & 0.052 & 0.619 & 0.454 & 0.231 & 0.043 \\
 & Qwen3-VL-32B (V) & 0.854 & 0.603 & 0.946 & 0.000 & 0.833 & 0.298 & 0.663 & 0.507 & 0.643 & 0.573 & 0.733 & 0.500 \\
  & Qwen2.5-14B (T) & 0.707 & 0.709 & 0.622 & 0.205 & 0.833 & 0.231 & 0.137 & 0.099 & 0.571 & 0.349 & 0.464 & 0.241 \\
  & TS-Reasoner-7B & 0.902 & 0.706 & 0.919 & 0.428 & 0.929 & 0.383 & 0.700 & 0.586 & 0.663 & 0.371 & 0.743 & 0.631 \\
  & TimeOmni-1-7B & 0.658 & 0.535 & 0.297 & 0.137 & 0.738 & 0.374 & 0.117 & 0.069 & 0.244 & 0.148 & 0.316 & 0.191 \\
  & ChatTS-14B & \textbf{0.927} & \textbf{0.874} & \textbf{0.973} & \textbf{0.849} & 0.857 & 0.511 & 0.895 & 0.805 & \textbf{0.905} & \textbf{0.782} & \textbf{0.889} & 0.788 \\
  & \textbf{Ours} & 0.829 & 0.864 & \textbf{0.973} & 0.515 & \textbf{0.976} & \textbf{0.622} & \textbf{0.900} & \textbf{0.813} & 0.737 & 0.778 & 0.864 & \textbf{0.805} \\
\midrule
\rowcolor{gray!15} B & GPT-4o (T) & 0.690 & 0.825 & 0.732 & 0.474 & 0.573 & 0.331 & 0.191 & 0.136 & 0.324 & 0.281 & 0.366 & 0.284 \\
\rowcolor{gray!15}   & GPT-4o (V) & 0.702 & 0.361 & 0.938 & 0.589 & 0.610 & 0.398 & 0.375 & 0.265 & 0.367 & 0.389 & 0.472 & 0.311 \\
\rowcolor{gray!15}   & GPT-5 (T) & 0.655 & 0.914 & 0.948 & 0.790 & 0.768 & 0.476 & 0.648 & 0.519 & 0.527 & 0.445 & 0.602 & 0.606 \\
\rowcolor{gray!15}   & GPT-5 (V) & 0.738 & 0.620 & 0.959 & 0.701 & 0.793 & 0.576 & 0.648 & 0.507 & 0.633 & 0.530 & 0.660 & 0.544 \\
\rowcolor{gray!15}   & GPT-5 (T+V) & 0.750 & 0.922 & 0.990 & 0.841 & 0.829 & 0.554 & 0.648 & 0.537 & 0.629 & 0.511 & 0.661 & 0.627 \\
  & Qwen3-VL-8B (V) & 0.167 & 0.022 & 0.155 & 0.093 & 0.146 & 0.015 & 0.074 & 0.041 & 0.400 & 0.373 & 0.255 & 0.042 \\
  & Qwen3-VL-32B (V) & 0.786 & 0.382 & 0.918 & 0.549 & 0.744 & 0.221 & 0.375 & 0.281 & 0.516 & 0.417 & 0.532 & 0.318 \\
  & Qwen2.5-14B (T) & 0.711 & 0.669 & 0.705 & 0.217 & 0.256 & 0.094 & 0.111 & 0.082 & 0.402 & 0.276 & 0.339 & 0.193 \\
  & TS-Reasoner-7B & 0.905 & 0.600 & 0.990 & 0.778 & 0.976 & 0.403 & 0.481 & 0.341 & 0.665 & 0.586 & 0.675 & 0.425 \\
  & TimeOmni-1-7B & 0.524 & 0.231 & 0.515 & 0.366 & 0.329 & 0.166 & 0.068 & 0.028 & 0.247 & 0.011 & 0.195 & 0.094 \\
  & ChatTS-14B & \textbf{0.976} & 0.902 & \textbf{1.000} & \textbf{0.930} & 0.927 & 0.572 & 0.828 & 0.752 & 0.818 & 0.834 & 0.862 & 0.787 \\
  & \textbf{Ours} & 0.964 & \textbf{0.906} & \textbf{1.000} & 0.920 & \textbf{0.988} & \textbf{0.705} & \textbf{0.917} & \textbf{0.860} & \textbf{0.829} & \textbf{0.846} & \textbf{0.893} & \textbf{0.868} \\
\bottomrule
\end{tabular}
\end{table*}

\begin{table}[tbp]\centering
\caption{ChatTS reasoning evaluation (Dataset A). Inductive reasoning is free-form Q\&A scored with the RAGAS pipeline; the other tasks are multiple-choice or true/false accuracy. MCQ2 samples 100 questions from the two-series set. Gray rows denote reference models. Bold values mark the best result per column, excluding reference models.}
\label{tab:chatts_reasoning_app}
\begin{tabular}{l cccc c}
\toprule
Model & Induct. & Deduct. & Causal & MCQ2 & Average \\
\midrule
\rowcolor{gray!15} GPT-4o (T)   & 0.336 & 0.628 & 0.685 & 0.510 & 0.540 \\
\rowcolor{gray!15} GPT-4o (V)   & 0.322 & 0.605 & 0.652 & 0.490 & 0.517 \\
\rowcolor{gray!15} GPT-5 (T)    & 0.277 & 0.581 & 0.663 & 0.730 & 0.563 \\
\rowcolor{gray!15} GPT-5 (V)    & 0.382 & 0.674 & 0.707 & 0.675 & 0.610 \\
\rowcolor{gray!15} GPT-5 (T+V)  & 0.315 & 0.605 & 0.717 & 0.760 & 0.599 \\
Qwen3-VL-8B (V)       & 0.197 & 0.558 & 0.685 & 0.480 & 0.480 \\
Qwen3-VL-32B (V)      & 0.399 & 0.651 & 0.750 & 0.610 & 0.603 \\
Qwen2.5-14B (T)       & 0.184 & 0.605 & 0.348 & 0.320 & 0.364 \\
ChatTS-14B            & \textbf{0.518} & \textbf{0.744} & 0.804 & 0.600 & \textbf{0.667} \\
TS-Reasoner-7B        & 0.282 & 0.632 & 0.717 & 0.610 & 0.560 \\
TimeOmni-1-7B         & 0.040 & 0.419 & 0.400 & 0.580 & 0.360 \\
\textbf{Ours}         & 0.479 & 0.535 & \textbf{0.826} & \textbf{0.620} & 0.615 \\
\bottomrule
\end{tabular}
\end{table}

\begin{table}[tbp]\centering
\caption{TSExam accuracy across the 746 questions, by category. Gray rows denote reference models. Bold values mark the best result per column, excluding reference models.}
\label{tab:tsexam_app}
\begin{tabular}{l c ccccc}
\toprule
Model & Overall & Anomaly & Causality & Noise & Pattern & Similarity \\
\midrule
\rowcolor{gray!15} GPT-4o (T)   & 0.586 & 0.491 & 0.458 & 0.667 & 0.594 & 0.667 \\
\rowcolor{gray!15} GPT-4o (V)   & 0.736 & 0.583 & 0.611 & 0.786 & 0.787 & 0.758 \\
\rowcolor{gray!15} GPT-5 (T)    & 0.724 & 0.704 & 0.556 & 0.679 & 0.782 & 0.700 \\
\rowcolor{gray!15} GPT-5 (V)    & 0.784 & 0.648 & 0.431 & 0.750 & 0.906 & 0.775 \\
\rowcolor{gray!15} GPT-5 (T+V)  & 0.808 & 0.731 & 0.417 & 0.738 & 0.931 & 0.792 \\
Qwen3-VL-8B (V)       & 0.618 & 0.546 & 0.514 & 0.643 & 0.630 & 0.692 \\
Qwen3-VL-32B (V)      & 0.713 & 0.639 & 0.431 & 0.690 & 0.793 & 0.725 \\
ChatTS-14B            & 0.570 & 0.481 & 0.500 & 0.619 & 0.575 & 0.642 \\
TS-Reasoner-7B        & 0.441 & 0.343 & 0.458 & 0.512 & 0.442 & 0.467 \\
TimeOmni-1-7B         & 0.483 & 0.389 & 0.375 & 0.583 & 0.497 & 0.517 \\
\textbf{Ours}         & \textbf{0.899} & \textbf{0.759} & \textbf{0.764} & \textbf{0.988} & \textbf{0.948} & \textbf{0.900} \\
\bottomrule
\end{tabular}
\end{table}

\section{Training Data Details}
\label{app:data}

This section provides the composition and sampling procedure of the data used in the two training stages. Stage I aligns the complementary visual representations with the language-model representation space using descriptive time-series supervision. Stage II performs instruction tuning over a mixture of time-series understanding, numerical, and reasoning tasks. We distinguish throughout between the number of examples available in each source (\emph{raw examples}) and the number sampled during one training epoch (\emph{effective draws}), since several sources are explicitly rebalanced.

\subsection{Training Corpus Overview}
\label{app:data_overview}

Table~\ref{tab:training_data_overview} summarizes the datasets used in both training stages. Stage I contains 55,696 raw examples drawn from three captioning sources. Because these sources differ substantially in size, we sample them with equal probability at the source level, such that each contributes one third of the effective Stage-I distribution. We draw 55,696 examples per epoch and train for three epochs.

Stage II combines instruction-following data, multiple-choice and numerical supervision, and a 100,000-example reasoning corpus. The effective training epoch contains 182,815 draws. The reasoning corpus and numerical corpus are retained at their natural sizes, contributing 100,000 and 11,000 examples, respectively. The remaining 71,815 draws are distributed approximately equally among ChatTS IFT, ChatTS SFT, and TSExam perception data. Consequently, the effective training distribution differs from the raw corpus distribution, preventing the largest individual source from dominating instruction tuning.

\begin{table*}[t]
\centering
\small
\setlength{\tabcolsep}{6pt}
\caption{
Training-data composition. Raw examples denote the number of available examples before sampling, whereas effective draws denote the approximate number presented in one training epoch. Stage-I sources are sampled with equal source probability. In Stage II, the reasoning and numerical corpora retain their natural sizes, while the remaining sampling budget is divided approximately equally among the three instruction/multiple-choice sources.
}
\label{tab:training_data_overview}
\begin{tabular}{llrrl}
\toprule
Stage & Source & Raw Examples & Effective Draws / Epoch & Supervision \\
\midrule
I & CaTS & 15,995 & $\sim$18,565 & Caption / description \\
I & ChatTS-style captions & 32,000 & $\sim$18,565 & Caption / description \\
I & Structural captions & 7,701 & $\sim$18,565 & Caption / description \\
\cmidrule(lr){2-5}
& \textit{Stage I total} & \textit{55,696} & \textit{55,696} & \\
\midrule
II & ChatTS IFT & 6,373 & $\sim$23,938 & Instruction following \\
II & ChatTS SFT & 44,802 & $\sim$23,938 & Instruction following \\
II & TSExam perception & 20,640 & $\sim$23,939 & Multiple choice \\
II & Numerical corpus & 11,000 & 11,000 & Numerical \\
II & Reasoning corpus & 100,000 & 100,000 & Reasoning / multiple choice \\
\cmidrule(lr){2-5}
& \textit{Stage II total} & \textit{182,815} & \textit{182,815} & \\
\bottomrule
\end{tabular}
\end{table*}

\subsection{Stage I: Modality-Alignment Data}
\label{app:stage1_data}

Stage I is designed to align the visual time-series representations with language while preserving the pretrained capabilities of the underlying vision-language model. Its supervision therefore consists of descriptive time-series examples rather than the downstream reasoning tasks used in Stage II.

The corpus combines 15,995 CaTS examples, 32,000 ChatTS-style caption examples, and 7,701 structural caption examples, for 55,696 raw examples in total. The substantial difference in source sizes motivates source-level balancing. At each draw, one of the three sources is selected with equal probability, after which an example is sampled from that source. Thus, each source accounts for approximately one third of the effective Stage-I training distribution, irrespective of its raw size. We use 55,696 draws per epoch for three epochs. This stage exposes the model to diverse descriptions of temporal signals and
provides supervision for associating the visual representations with their
corresponding temporal semantics before instruction tuning.

\subsection{Stage II: Instruction-Tuning Data}
\label{app:stage2_data}

Stage II broadens the supervision from descriptive alignment to instruction-conditioned time-series understanding and reasoning. The training mixture consists of ChatTS instruction-following data, TSExam perception questions, the numerical corpus, and a separately constructed reasoning corpus.

The raw instruction-tuning sources contain 6,373 ChatTS IFT examples, 44,802 ChatTS SFT examples, 20,640 TSExam perception examples, 11,000 numerical corpus examples, and 100,000 reasoning examples. Rather than sampling this mixture according to raw source frequency, we explicitly control its effective composition. The 100,000 reasoning examples and 11,000 numerical examples are retained in each epoch. The remaining 71,815 draws are divided approximately
equally among ChatTS IFT, ChatTS SFT, and TSExam perception data, yielding approximately 23.9k draws from each source and 182,815 total draws per epoch.

The reasoning corpus is designed to cover a broad range of temporal reasoning capabilities and interaction formats, including numerical reasoning, event
prediction, entity recognition, causal discovery, sequence continuation, dynamical reasoning, temporal reasoning, abductive reasoning, and inductive
reasoning. While these capabilities overlap with those evaluated by TSRBench, the underlying time series, questions, and supervision in the reasoning corpus are constructed independently.

\paragraph{Evaluation-set separation.}
No evaluation examples from TSRBench, TSExam-HF, or the ChatTS evaluation sets are included in the training mixture. This separation applies to both training stages.

\subsection{Reasoning Corpus Specification}
\label{app:reasoning_corpus}

The Stage-II reasoning corpus contains 100,000 examples spanning nine task
families and 25 generation templates. Table~\ref{tab:reasoning_corpus}
summarizes the size of each family, the underlying data-generating process,
the number of input channels and answer choices, and the supervision format.
The task families are denoted NR (numerical reasoning), EP (event prediction),
ER (entity recognition), CD (causal discovery), TSF (time-series
forecasting), DR (dynamical reasoning), TR (temporal reasoning), AR
(abductive reasoning), and IR (inductive reasoning).

The corpus deliberately combines tasks with different forms of supervision.
NR, EP, and IR require explicit numerical computation or intermediate
multi-step inference and together account for 36,328 examples. For these
families, we retain rationale supervision followed by the final answer.
The remaining six families---CD, ER, TSF, AR, DR, and TR---account for 63,672
examples and are trained using the final option letter directly. Although
deterministically generated rationales are available for 97,844 of the
100,000 examples, we discard these rationales for the latter families in the
final training configuration rather than applying chain-of-thought
supervision uniformly across all tasks.

\begin{table*}[t]
\centering
\small
\setlength{\tabcolsep}{3pt}
\caption{
Taxonomy of the reasoning corpus (100,000 examples across
25 templates). Shares are computed over the complete reasoning corpus. Where
a family mixes sources, \emph{Channels} and \emph{Options} are listed per
source in the order of the description.
}
\label{tab:reasoning_corpus}
\begin{tabularx}{\textwidth}{@{}l c r r >{\raggedright\arraybackslash}X c c >{\raggedright\arraybackslash}p{1.55cm}@{}}
\toprule
Family & Tmpl. & Items & Share & Underlying Generation / Governing Process & Channels & Options & Output \\
\midrule
NR & 8 & 16,874 & 16.9\% & Eight physical formulas: traffic flow, cooling load, growing degree days, seismic S--P distance, reaction kinetics, Penman--Monteith evapotranspiration, intersection delay, and wind power. & 2--4 & 4 & Rationale + Letter \\
EP & 1 & 15,225 & 15.2\% & Five-channel synthetic weather traces in which rainfall events are triggered by a latent humidity surge and barometric-pressure drop; two informative and three nuisance channels. & 5 & 2 & Rationale + Letter \\
ER & 2 & 14,802 & 14.8\% & Parameterized light curves spanning five variable-star classes and tri-axial accelerometer segments spanning six physical-activity classes. & 1 / 3 & 5 / 6 & Option Letter \\
CD & 1 & 12,687 & 12.7\% & Random five-node directed acyclic graphs simulating sensor networks across heterogeneous ranges; the task is to identify the generating adjacency structure. & 5 & 4 & Option Letter \\
TSF & 1 & 12,500 & 12.5\% & Histories containing trend, seasonality, and autoregressive structure; the task is to select the appropriate continuation trajectory among generated alternatives. & 4 & 4 & Option Letter \\
DR & 3 & 10,573 & 10.6\% & Propagation of dynamical processes including AR(1), geometric growth, and continuous Lorenz-attractor state trajectories. & 4 & 4 & Option Letter \\
TR & 2 & 6,766 & 6.8\% & Synthetic conflict--cooperation proxy series paired with timestamped event summaries; tasks evaluate event ordering and before/after precedence. & 1 & 4 & Option Letter \\
AR & 5 & 6,344 & 6.3\% & Real NBA and NFL play-by-play games with win-probability trajectories and a masked interval of ten plays; the task is to select the first hidden play among four candidates (2,156 items). Synthetic trend-plus-season series with one injected anomaly (spike, cutoff, scale, speed-up, sign-flip, or wander); the task is to name the anomaly class or the index band containing it (4,188 items). & 2 / 1 & 4 / 5 & Option Letter \\
IR & 2 & 4,229 & 4.2\% & Long-range temporal patterns based on monthly sunspot sequences and multichannel climate indicators used to infer higher-level temporal outcomes. & 1 / 8 & 4 & Rationale + Letter \\
\bottomrule
\end{tabularx}
\end{table*}

\paragraph{Data provenance.}
The reasoning corpus combines procedurally generated time series with tasks
constructed from real underlying temporal data. Most families use synthetic
generative processes that provide direct control over the latent mechanism and
the correct answer. The only real temporal data are NBA (2021--22) and NFL (2021--22 to
2023--24) play-by-play records from GAMETime~\citep{tan2025inferring}, used in two AR templates (2,156 items, 2.2\%
of the corpus); for these, the questions and supervision are constructed
independently of the evaluation sets.
This design allows the corpus to combine controlled reasoning problems with
temporal patterns occurring in real data without using examples from the
evaluation sets described above.

\subsection{Numerical Corpus Specification}
\label{app:numeric_corpus}

The Stage-II numerical corpus contains 11,000 examples spanning three
subcategories and 20 generation templates. Table~\ref{tab:numeric_corpus}
summarizes the size of each subcategory and the statistics it targets. Every
item is a single series of length 256 drawn from a shared pool of eleven
generating recipes (trends, periodic waves, noise, and autoregressive and
moving-average processes) with additive observation noise. The target is the
requested statistic computed exactly on the observed series and rounded to two
decimals; there are no answer options.

\begin{table*}[t]
\centering
\small
\setlength{\tabcolsep}{3pt}
\caption{
Taxonomy of the numerical corpus (11,000 examples across 20 templates).
Shares are computed over the complete numerical corpus.
}
\label{tab:numeric_corpus}
\begin{tabularx}{\textwidth}{@{}l c r r >{\raggedright\arraybackslash}X c c >{\raggedright\arraybackslash}p{1.55cm}@{}}
\toprule
Subcategory & Tmpl. & Items & Share & Target Statistic & Channels & Options & Output \\
\midrule
Statistical retrieval & 10 & 5,500 & 50.0\% & Maximum, minimum, mean, median, first and third quartile, standard deviation, variance, first value, and last value. & 1 & --- & Scalar \\
Signal properties & 5 & 2,750 & 25.0\% & Range, interquartile range, mean absolute value, lag-1 autocorrelation, and number of zero crossings. & 1 & --- & Scalar \\
Step dynamics & 5 & 2,750 & 25.0\% & Mean absolute first difference, largest one-step jump, best-fit slope, first-half minus second-half mean, and energy. & 1 & --- & Scalar \\
\bottomrule
\end{tabularx}
\end{table*}

\end{document}